\documentclass{article}

\usepackage{arxiv}

\usepackage[utf8]{inputenc}
\usepackage[T1]{fontenc}
\usepackage{lmodern}
\usepackage{microtype}

\usepackage{amsmath,amssymb,amsfonts}
\usepackage{graphicx}
\usepackage{bm}
\usepackage{booktabs}
\usepackage{subcaption}
\usepackage{enumitem}
\usepackage{multirow}
\usepackage{tikz}
\usepackage{algorithm}
\usepackage{algorithmic}
\usepackage{nicefrac}

\usepackage{hyperref}
\usepackage{url}
\usepackage{doi}

\usepackage[authoryear,longnamesfirst]{natbib}

\title{A comparative study of accuracy and rollout stability of temporal surrogate models}

\author{
  Rajarshi Biswas\thanks{Corresponding author. Email: \texttt{bis.rajarshi@gmail.com}} \\
  Senior AI/Data Scientist \\
  Cargill Inc. \\
  Wayzata, MN 55391, USA \\
  bis.rajarshi@gmail.com
}

\date{}

\renewcommand{\headeright}{}
\renewcommand{\undertitle}{}
\renewcommand{\shorttitle}{Temporal Surrogates Comparison}

\begin{document}

\maketitle

\begin{abstract}
Temporal surrogate models are effective for predicting chaotic dynamical systems where computational cost can be prohibitive. Several deep neural network architectures can be used for such purposes. In this work, a few commonly used architectures are compared using a common training protocol. The objective is to fairly assess the impact of model architectures for long-horizon prediction stability. Experiments are carried out for three problems, the double pendulum, the Kuramoto-Sivashinsky equations, and the Kolmogorov flow. The experiments are carried out with matching model capacity. Analysis is also carried out for a scenario where each model is individually optimized.  It is observed that in both scenarios, the models exhibit categorical differences in long-horizon rollouts. For a concrete quantification, stepwise error injections and perturbation amplifications are analyzed using metrics such as local jacobian, relative one-step bias, and finite-time Lyapunov growth. Additionally, an attractor analysis is also conducted to assess how well the learned models replicate the underlying system geometry. An ablation study to isolate the impact of each component of a continuous-update architecture is also carried out. It is concluded that models that having integrator-like updates show lower bias and perturbation amplification yielding stable long-horizon rollout and more accurate predictions.
\end{abstract}

\keywords{Nonlinear dynamics \and Deep neural networks \and Chaotic systems \and Surrogate modeling}

% Main text
\section{Introduction}\label{sec:intro}
For many engineering applications, such as turbulent fluid flow, multi-phase flow, weather predictions, etc., accurate modeling and prediction of chaotic dynamical systems are often a critical requirement. From low-dimensional mechanical to large-scale industrial machines, almost all systems are built on components that exhibit non-linear or multiscale behavior. Traditionally, system designers have leveraged high-fidelity numerical solvers to model the dynamics of such components and make predictions of the system's evolution. However, as the degree of freedom increases, so do computational expenses, often becoming a bottleneck for problems of scale. Scientific machine learning-based methods attempt to alleviate such bottlenecks and have been very successful in recent years.

With the availability of large volumes of data, data-driven approaches have become a popular choice for predicting systems governed by underlying physical laws. Data-driven models, such as multilayer perceptrons (MLP), recurrent neural networks (RNN), long short-term memory networks (LSTM) have been used in previous studies ~\citep{vlachas2018chaoticrnn, pathak2018reservoir, lusch2018deep, geneva2020modeling, chattopadhyay2020lorenzlstm}. These models are designed to learn the temporal mapping from state $\mathbf{x}_t$ to $\mathbf{x}_{t+1}$ from training data. They typically show strong predictive accuracy in a short-horizon predictive setting. However, this modeling approach usually lacks the necessary inductive bias to capture the governing physics. As a result, in long-horizon configurations, particularly when deployed as a closed-loop rollout, the stepwise errors accumulate, leading to either grossly incorrect or completely divergent predictions.

Recently multiple research efforts have made progress in addressing the limitations mentioned above by proposing novel algorithms. For example, reservoir computing methods~\citep{pathak2018reservoir} have demonstrated highly accurate long-horizon predictions for chaotic turbulent flows. Similarly, Koopman operator-based approaches~\citep{lusch2018deep, brunton2022koopman} that linearly approximate the evolution of nonlinear dynamics in encoded latent spaces have shown great potential. Direct embedding of physical systems into neural network architectures is another strong approach to introduce inductive biases into the learned models. To that end, several novel architectures can be identified. For example, Hamiltonian and Lagrangian neural networks encode conservation laws in the model~\citep{greydanus2019hamiltonian, cranmer2020lagrangian}, whereas symplectic integrator networks and geometric deep learning methods encode structural or geometrical constraints~\citep{zhong2019symplecticodenet, jin2020sympnets, finzi2020lagham}. Physics-informed neural network~\citep{raissi2019pinn} is another interesting framework that uses residuals of the governing equations to guide the models learn the underlying governing equations more accurately. Operator-learning approaches such as DeepONet~\citep{lu2021deeponet} and Fourier Neural Networks(FNO)~\citep{li2021fourier} learn function space mappings and have demonstrated remarkable predictive capabilities.

A class of sophisticated model architecture has also been used for chaotic predictions. For example, Neural ordinary differential equations (NeuralODE) by Chen et al.,~\citep{chen2018neuralode}  model system dynamics as continuous time-vector fields integrated with numerical solvers. Another work by Bai et al.,~\citep{bai2018tcn} uses Temporal convolutional networks (TCN) to leverage large receptive fields via causal dilated convolutions have been used for chaotic forecasting. It is therefore fair to conclude that there currently exists a plethora of highly advanced neural network architectures that can be leveraged for chaotic forecasting. However, as the complexity of the models grows, it becomes increasingly important to understand the underlying mechanisms. Controlled experiments provide a reliable and effective way to compare different architectures under homogeneous conditions. Despite this, a systematic comparison between discrete-time and continuous-time autoregressive mechanisms in a controlled experiment configuration remains limited.

In the present work, such a controlled experiment is undertaken where five architectures, namely: MLP, LSTM, TCN, NeuralODE, and a residual update-style Conditioned Residual Dynamics (CoRD), are examined and compared thoroughly. The first three models can be placed under the discrete-time update category, whereas the latter two fall under the continuous update category. These model architectures are well-suited for a capacity-matched experiment where the network depth is held constant, while the width is carefully chosen to match the total hyperparameter count within a tight tolerance. For the continuous update models, the one-step map $\mathbf{x}_t \mapsto \mathbf{x}_{t+1}$ is expressed as a residual update,

\begin{equation}
    \mathbf{x}_{t+1} = \mathbf{x}_t + \Delta t \, f_\theta([\mathbf{x}_t, \mathbf{x}_0]),
\end{equation}

NeuralODE integrates the continuous-time system $d\mathbf{x}/dt = f_\theta([\mathbf{x},\mathbf{x}_0])$ using an adaptive solver, whereas CoRD implements this update using fixed-step explicit Euler substeps. The underlying framework for both NeuralODE and CoRD matches the MLP version. During model training, a one-step teacher-forcing protocol is leveraged. In this, the models are conditioned on the initial trajectory state and trained to predict the next time state ~\citep{williams1989learning}. Although more improved multi-step approaches ~\citep{bengio2015scheduled, lamb2016professor} exist that bypass the exposure-bias issue of teacher-forcing, such methods introduce additional architectural interactions and could confound the experiment. A uniform one-step objective is intentionally chosen here to isolate the effect of temporal update structure under identical conditions. 

Three benchmarks are chosen for the current study in order of increasing dynamic complexity. The first is the double pendulum problem, which represents low-dimensional Hamiltonian chaos. Next is the Kuramoto-Sivashinsky equation (KS Equation) ~\citep{sivashinsky1977, kuramoto1978diffusion}, which exhibits one-dimensional dissipative spatio-temporal chaos. Finally, a two-dimensional Kolmogorov flow ~\citep{lucas2014kolmogorov} is used because it models forced Navier-Stokes dynamics with sustained chaotic vortical structures. These problems represent a progression from low to high dimensional chaotic motion. The models are trained in the state space for the double pendulum. However, for the PDE driven problems, autoencoders are used to obtain latent spaces, which are used by the models to learn the system dynamics.

Assessments are carried out on the long-horizon closed-loop rollout predictions from each of the models for each benchmark. In addition to comparing aggregate trajectory error metrics, regime-dependent analysis is carried out to assess the impact of the initial system state. This is done by stratifying the trajectories by initial total energy for the double pendulum, spatial variance for KS Equations, and enstrophy for Kolmogorov flow. Emphasis is given to understanding the key drivers of error growth over time. For this purpose, two complementary mechanisms are devised: a) one-step bias injection, and b) multi-step perturbation analysis. These are quantified via a local Jacobian diagnosis, relative bias measures, and finite-time perturbation metrics. The models' capabilities to represent the attractor geometry are quantified via invariant statistical measurements and principal-component projections of the reconstructed fields for KS Equations and Kolmogorov flow. Finally, an architecture ablation study is carried out on the KS Equations to isolate the effects of individual components of the residual update model architecture. In this, the individual contributions of design strategies such as residual updates, temporal sub-stepping, global conditioning, etc., are examined. 

This work presents a detailed understanding of temporal update mechanisms especially for stability and long-horizon behavior for chaotic surrogate modeling via controlled experiments. The experiments:
a) use a common dataset with identical normalization, training, and inference setup,
b) use capacity-matched configurations to isolate the effect of architectural structure,
c) quantify the relation of short-horizon prediction accuracy to long-time stability and attractor fidelity.

Across all the benchmarks considered here, temporal update structure is observed to have a wider impact on the predictive accuracy and stability than the total parameter count. These findings provide an empirical characterization of the behavior of common temporal models.

The remainder of the paper is organized as follows. Section~\ref{sec:methods} provides details of the architectures used and explains the homogeneous training \& inference protocol. Section~\ref{sec:Experiments} presents the canonical problems, training data generation strategy for each, and summarizes the model parameter counts. Section~\ref{sec:results} discusses aggregate predictive performances. A detailed dynamical analysis to quantify one-step error, perturbation growth and adherence to attractor geometry via metrics such as spectral radii of the one-step jacobian, relative error injection, and finite time Lyapunov exponent, is presented in Section~\ref{sec:analysis}. Section~\ref{sec:arch_ablation} is an ablation study for a deeper understanding of the architecture components. Finally, Section~\ref{sec:discussion} summarizes the key findings and takeaways.

%##################################################################################
\section{Methods}
\label{sec:methods}
%##################################################################################
This section describes the modeling framework used for the experiments. Details of the neural network architectures along with the training and inference protocol are also covered. Hyperparameters are chosen such that model capacities are approximately matched while keeping the model depths consistent (Section~\ref{subsec:hyperparams}).
%============================================
\subsection{Problem Formulation and Notation}
\label{subsec:problem_formulation}
\begin{figure}
\centering
\includegraphics[width=0.8\linewidth]{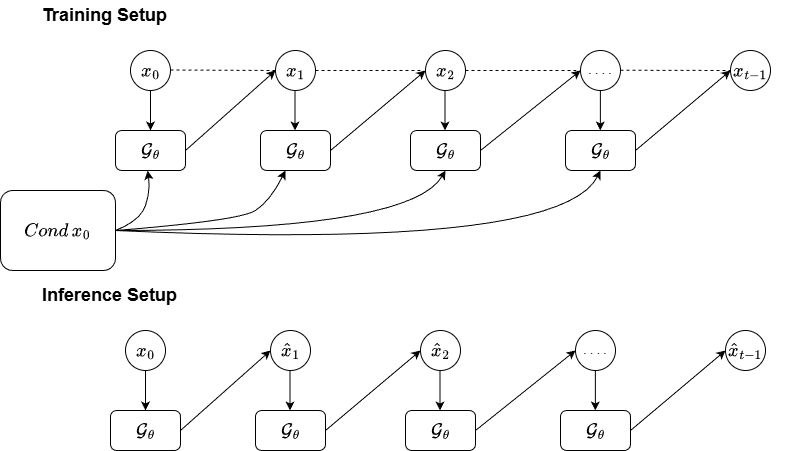}
\caption{\textit{Training and inference setup for the autoregressive temporal models with initial-state conditioning.}}
\label{fig:temporal_model_schematic}
\end{figure}

%============================================
Datasets of discrete-time trajectories can be presented as,
\begin{equation}
    \{\mathbf{x}_t^{(n)}\}_{t=0}^{T-1}, \qquad \mathbf{x}_t^{(n)} \in \mathbb{R}^D,    
\end{equation}
where $n$ is the trajectory index and $t$ represents time steps. The vector $\mathbf{x}_t$ denotes the model state. For low-dimensional ODE benchmarks, $\mathbf{x}_t$ coincides with the physical state. For PDE benchmarks, the physical field is first encoded using a pre-trained autoencoder, producing latent states $\mathbf{z}_t$. To unify the notation across experiments, the state used by the temporal models is denoted by $s_t$, where,
\begin{equation}
    s_t =
        \begin{cases}
        \mathbf{x}_t, & \text{ODE benchmarks}, \\
        \mathbf{z}_t, & \text{latent PDE benchmarks}.
        \end{cases}    
\end{equation}
All states are standardized componentwise using statistics computed from the training split,
\begin{equation}
\tilde{s}_t = \frac{s_t - \mu}{\sigma},
\label{eq:normalization}
\end{equation}
where $\mu$ and $\sigma$ denote the training-set mean and standard deviation.
The temporal models learn a one-step mapping
\begin{equation}
    G_\theta : (\tilde{s}_t,\tilde{s}_0) \rightarrow \tilde{s}_{t+1},    
\end{equation}
where the current state is augmented by the initial state of the trajectory,
\begin{equation}
    \mathbf{u}_t = [\tilde{s}_t,\tilde{s}_0] \in \mathbb{R}^{2D}.    
\end{equation}
This conditioning provides global trajectory context throughout the rollout.
Training uses a one-step teacher-forcing objective, while evaluation is performed through closed-loop autoregressive rollouts. Figure~\ref{fig:temporal_model_schematic} illustrates this setup.
%============================================
\subsection{Multilayer Perceptron (MLP)}
\label{subsec:autoreg_mlp}
%============================================
The first architecture is a standard feed-forward multilayer perceptron that predicts the next normalized state,
\begin{equation}
    \tilde{s}_{t+1,\mathrm{pred}} = f_{\mathrm{MLP}}(\mathbf{u}_t).    
\end{equation}
The network consists of fully connected layers with GELU nonlinear mappings. Further details are covered in ~\ref{app:mlp_arch}. Because the MLP is memoryless, temporal error propagation occurs only through the fed-back state during rollout.
%============================================
\subsection{Long Short-Term Memory (LSTM)}
\label{subsec:autoreg_lstm}
%============================================
The second architecture is a recurrent LSTM network,
\begin{equation}
    (\mathbf{h}_{t+1},\mathbf{c}_{t+1}) = \mathrm{LSTM}(\mathbf{u}_t,\mathbf{h}_t,\mathbf{c}_t),    
\end{equation}
followed by the readout
\begin{equation}
    \tilde{s}_{t+1,\mathrm{pred}} = \mathbf{W}_{\mathrm{out}}\mathbf{h}_{t+1}+\mathbf{b}_{\mathrm{out}}.    
\end{equation}
Hidden and cell states are initialized to zero at the start of each rollout. Internal gating equations are detailed in ~\ref{app:lstm_arch}. It is important to note that, LSTM parameters are distributed across multiple gates. Therefore a fixed parameter budget yields a smaller effective hidden width than an equivalently sized MLP.
%============================================
\subsection{Temporal Convolutional Network (TCN)}
\label{subsec:autoreg_tcn}
%============================================
The TCN architecture uses stacks of dilated causal convolutions applied to the sequence of conditioned inputs. Given $(\mathbf{u}_0,\dots,\mathbf{u}_{T-2})$, the network computes
\begin{equation}
    \mathbf{h}_{0:T-2} = \mathrm{TCN}(\mathbf{u}_{0:T-2})
\end{equation}
followed by
\begin{equation}
    \tilde{s}_{t+1,\mathrm{pred}} = \mathbf{W}_{\mathrm{out}}\mathbf{h}_t+\mathbf{b}_{\mathrm{out}}    
\end{equation}
Causal convolutions ensure that predictions depend only on current and past inputs without any leakage from future states. Increasing the dilation  factor expands the temporal receptive field allowing for a longer temporal context without proportional increase in the parameter count. Further architectural details are provided in ~\ref{app:tcn_arch}.
%============================================
\subsection{Neural Ordinary Differential Equation (NeuralODE)}
\label{subsec:autoreg_neuralode}
%============================================
The NeuralODE architecture evolves the temporal dynamics through a continuous-time vector field,
\begin{equation}
    \frac{d\tilde{s}(s)}{ds} = f_\theta([\tilde{s}(s),\tilde{s}_0])
\label{eq:cont_intg}
\end{equation}
Starting from $\tilde{s}(0)=\tilde{s}_t$, the next prediction is obtained by integrating the ODE over a fixed interval $h_{\mathrm{ODE}}$,
\begin{equation}
    \tilde{s}_{t+1,\mathrm{pred}} = \tilde{s}(h_{\mathrm{ODE}})    
\end{equation}
For the current experiments, integration is performed using an adaptive Dormand-Prince solver~\citep{dormand1980family}. The vector field $f_\theta$ uses a MLP style backbone. This allows the effect of continuous-time integration to be isolated. ~\ref{app:neuralode_arch} covers more details on this architecture.
%============================================
\subsection{Conditioned Residual Dynamics Network (CoRD)}
\label{subsec:autoreg_CoRD}
%============================================
The CoRD model represents dynamics using a residual update,
\begin{equation}
\tilde{s}_{t+1,\mathrm{pred}} = \tilde{s}_t + \Delta t\, f_\theta(\mathbf{u}_t),
\label{eq:CoRD_residual_update_main}
\end{equation}
where $f_\theta$ is an MLP just like the NeuralODE. $\Delta t$ is the sampling interval of the dataset. This corresponds to an explicit Euler discretization of Eqn.~\ref{eq:cont_intg}. The choice of Euler substeps dictates numerical stability and computational expenses. For the current experiments, Euler substep is set to 3. ~\ref{app:cord_arch} has additional information on this architecture.
%============================================
\subsection{Training \& Inference Protocol}
\label{subsec:training_protocol}
%============================================
All models are trained using the Adam optimizer \citep{kingma2015adam} with a learning rate of $10^{-3}$. The training objective is the mean-squared one-step prediction error,
\begin{equation}
\label{eq:training_loss_eqn}
\mathcal{L}
=
\frac{1}{B(T-1)}
\sum_{b=1}^{B}
\sum_{t=0}^{T-2}
\|
\tilde{s}^{(b)}_{t+1,\mathrm{pred}}
-
\tilde{s}^{(b)}_{t+1}
\|_2^2.
\end{equation}
The gradients are clipped to unit $\ell_2$ norm to improve training stability. Checkpoints are used during training, the one with lowest validation loss is selected for evaluation.
Predictions are obtained using closed-loop autoregressive rollouts. During evaluation gradients are disabled. Given an initial state $s_0$, the normalized state $\tilde{s}_0$ is computed using the training statistics. Predictions are then generated recursively,
\begin{equation}
\hat{\tilde{s}}_{t+1}
=
G_\theta(\hat{\tilde{s}}_t,\tilde{s}_0),
\qquad
\hat{\tilde{s}}_0=\tilde{s}_0.    
\end{equation}
For latent PDE benchmarks, the physical field is first normalized and encoded to a latent state before applying the temporal model. The autoencoder weights are kept fixed during rollout. The predicted latent space is transformed back to the physical space using a frozen decoder.
%##################################################################################
\section{Experiments}
\label{sec:Experiments}
%##################################################################################
This section describes the benchmark systems used to evaluate the temporal models, including dataset generation, state representations, and evaluation protocols. Hyperparameters and dataset statistics are summarized in Table~\ref{tab:hyperparams}. All models follow the common training and inference procedures described in Section~\ref{sec:methods}.
%============================================
\subsection{Double Pendulum}
\label{subsec:dp}
%============================================
%---------------------
\subsubsection{Problem and Dataset}
\label{subsubsec:dp_state}
%---------------------
The planar double pendulum system consists of two unit masses connected by rigid links of unit lengths ($m_1=m_2=1$, $L_1=L_2=1$). This is a classic example of low-dimensional systems that exhibits chaotic behavior depending on the initial condition. Minor perturbations in the initial conditions can lead to rapidly diverging trajectories. This system has been extensively studied in literature and dynamics is very well understood. Therefore, this naturally fits as the first use case for this study. The physical state is defined by the joint angles $(\theta_1,\theta_2)$ and angular velocities $(\omega_1,\omega_2)$. To avoid discontinuities at the boundaries $\pm\pi$, a sin-cos transformation is used for the joint angles, thereby yielding 6 state-space variables as given by Eqn.~\ref{eq:dp_sincos},
\begin{equation}
\mathbf{x}_t =
(\sin\theta_1,\cos\theta_1,\sin\theta_2,\cos\theta_2,\omega_1,\omega_2)
\in \mathbb{R}^6 .
\label{eq:dp_sincos}
\end{equation}
Physical angles are recovered using $\theta_i=\operatorname{atan2}(\sin\theta_i,\cos\theta_i)$ with temporal unwrapping when computing physical-space errors.

Trajectories are generated by integrating the equations of motion using a fourth-order Runge–Kutta scheme over $T=10\,\mathrm{s}$. States are saved every $\Delta t=0.05\,\mathrm{s}$, with each step computed using $10$ internal sub-steps ($\Delta t_{\mathrm{fine}}=0.005\,\mathrm{s}$), producing $N_t=201$ snapshots per trajectory.
Initial conditions are sampled uniformly
\[
\theta_1,\theta_2 \sim \mathcal{U}[-\pi,\pi],
\qquad
\omega_1,\omega_2 \sim \mathcal{U}[-6,6]\;\mathrm{rad/s}.
\]
Trajectories exhibiting numerical instability are rejected and resampled. This sampling covers both oscillatory and high-energy rotational regimes.
The dataset contains $N=2000$ trajectories with a training to validation split of 0.8.
%---------------------
\subsubsection{Evaluation Metrics}
\label{subsubsec:dp_loss}
%---------------------
To quantify the prediction accuracy, mean squared error (Eqn.~\ref{eq:dp_mse}) is used. In addition to the full horizon, this metric is also calculated for an \textit{early} horizon defined by the first 0-2s of the trajectory evolution. This isolates the short-term accuracy from long-horizon chaotic divergence.
\begin{equation}
\mathrm{MSE}_{\mathrm{state}}
=
\frac{1}{N_t}
\sum_{t=0}^{N_t-1}
\|
\mathbf{s}^{\mathrm{pred}}_t
-
\mathbf{s}^{\mathrm{true}}_t
\|_2^2 
\label{eq:dp_mse}
\end{equation}
%============================================
\subsection{Kuramoto--Sivashinsky (KS) Equation}
\label{subsec:ks_experiment}
%============================================
%---------------------
\subsubsection{Problem and Dataset}
\label{subsubsec:ks_setup}
%---------------------
The second benchmark is the one-dimensional Kuramoto–Sivashinsky equation
\begin{equation}
u_t + u\,u_x + u_{xx} + u_{xxxx} = 0,
\qquad x \in [0,L),
\end{equation}
with periodic boundary conditions. This system exhibits the evolution of spatio-temporal chaos governed by non-linear advection, diffusion, and higher order dissipation. For large domains ($L=22$ used in the current study), this system displays a weakly turbulent regime and has been used in previous studies~\citep{kassam2005etdrk4}. KS equation system perfectly fits the intermediate regime between a low-dimensional system and a more complex turbulent flow like system, and is therefore chosen as the next benchmark problem.

To generate the dataset, the spatial field is discretized on $N_x=128$ grid points and integrated using a pseudospectral ETDRK4 solver with time step $\Delta t_{\mathrm{int}}=0.25$. Initial conditions are generated as low-pass filtered random Fourier fields with varying amplitudes and bandwidths to span low, medium, and high-energy regimes. Each regime contributes $200$ trajectories, yielding a data set of length 600 with the same training to validation split as before.
After a transient of $t_{\mathrm{trans}}=50$, snapshots are recorded every three solver steps, resulting in an effective model step,
\[
\Delta t_{\mathrm{model}}=0.75 .
\]
%---------------------
\subsubsection{Latent Autoencoder Representation}
\label{subsubsec:ks_latent}
\begin{figure}[]
\centering
\begin{subfigure}{.48\linewidth}
\includegraphics[width=\linewidth]{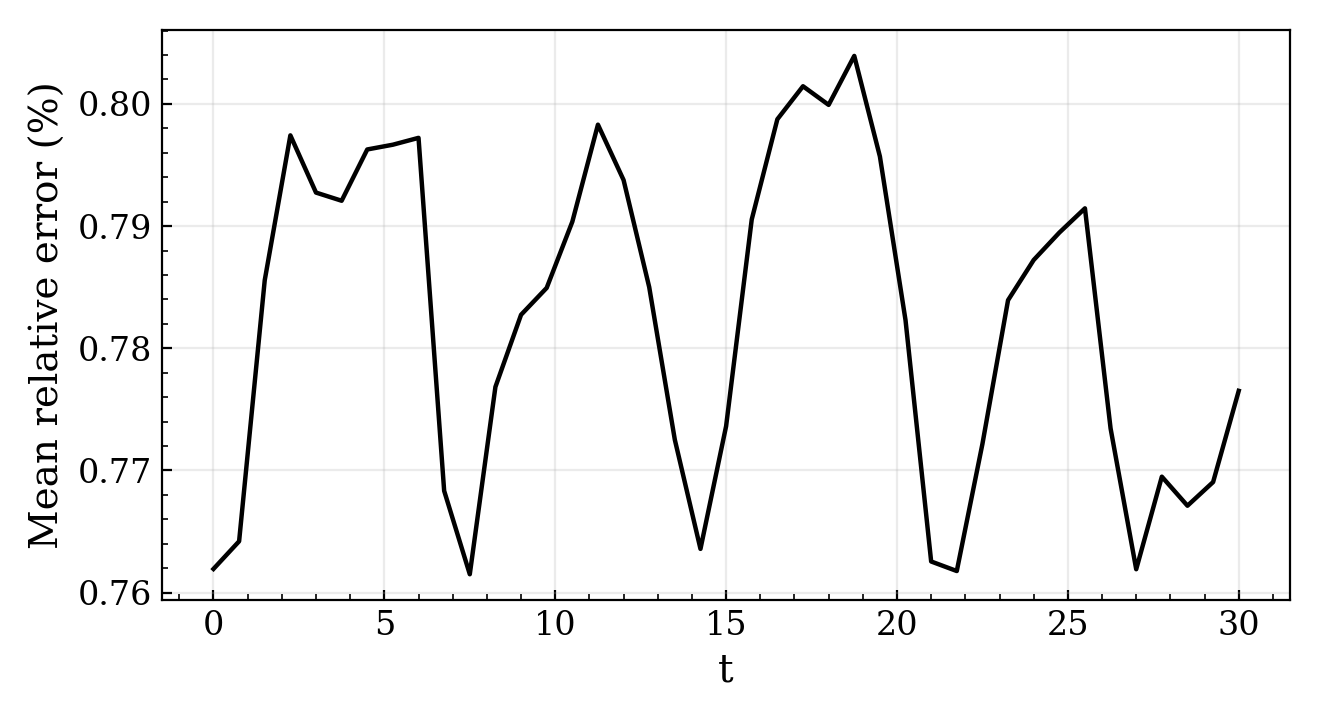}
\end{subfigure}
\hfill
\begin{subfigure}{.48\linewidth}
\includegraphics[width=\linewidth]{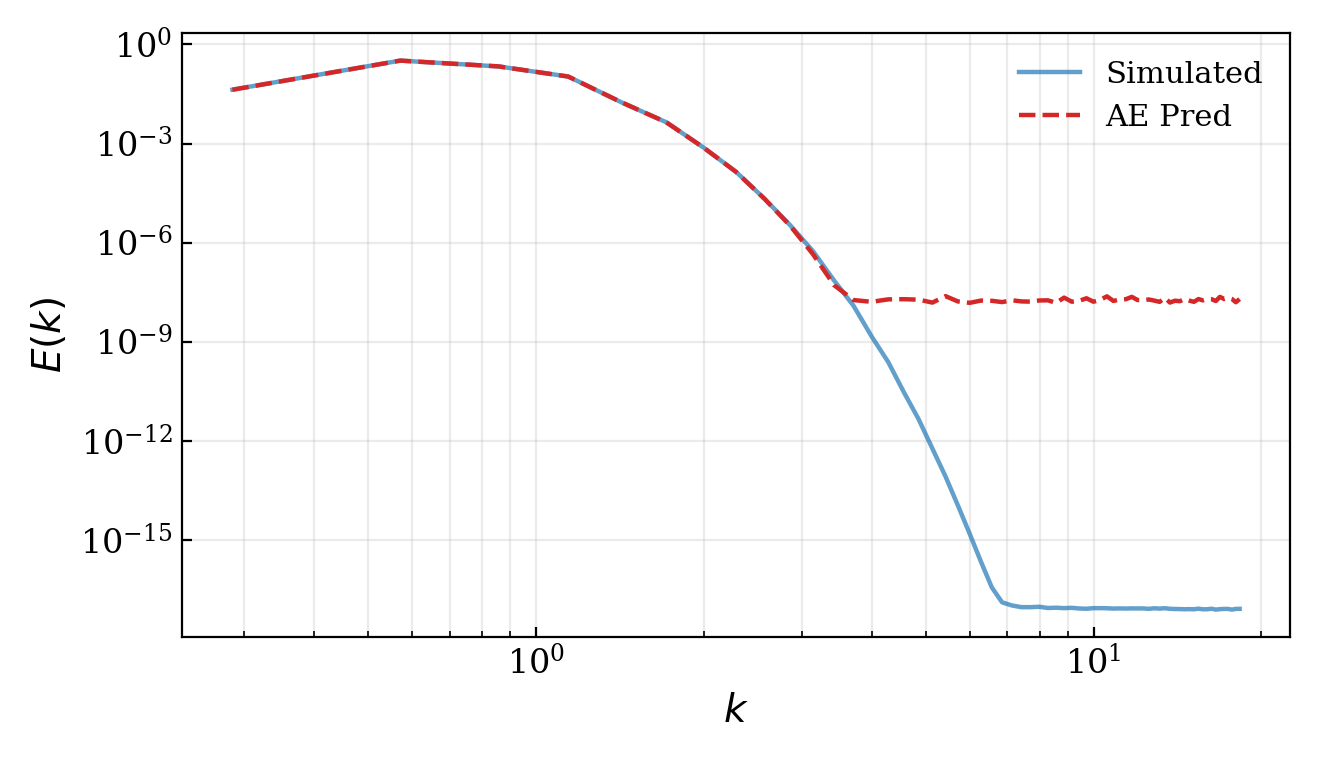}
\end{subfigure}
\caption{\textit{(Left) Mean relative reconstruction error over time on validation trajectories. (Right) Energy spectra of raw data and autoencoder reconstructions.}}
\label{fig:ks_ae_stats}
\end{figure}

%---------------------
Approximating a high-dimensional system such as the KS Equation requires significantly large neural architectures. To bypass this issue, autoencoders are leveraged that project the system into a lower dimensional latent space. The temporal evolution of this space is then modeled using the architectures as outlined above. Each spatial snapshot $u_t \in \mathbb{R}^{N_x}$ is encoded into a latent vector
\begin{equation}
\mathbf{z}_t=\mathcal{E}(u_t),
\qquad
\hat{u}_t=\mathcal{D}(\mathbf{z}_t)    
\end{equation}
using a fully connected autoencoder with GELU activations. The autoencoder is trained independently using reconstruction MSE and remains frozen across the architecture evaluations. The reconstruction quality is assessed on the validation set with the criteria that mean relative reconstruction error remains below $1\%$ across time. Additionally, it is ensured that the reconstructed fields preserve the spatial energy spectrum of the KS flow across dynamically active wavenumbers. Figure~\ref{fig:ks_ae_stats} demonstrates these criteria.
%---------------------
\subsubsection{Evaluation Metrics}
\label{subsubsec:ks_loss_eval}
%---------------------
Trajectory-level space–time MSE is used as the primary criteria for gauging predictive performances as given in Eqn.~\ref{eq:ks_mse},
\begin{equation}
\mathrm{MSE}^{(i)}_{\mathrm{traj}}
=
\frac{1}{N_t N_x}
\sum_{n=0}^{N_t-1}
\sum_{j=0}^{N_x-1}
(\hat{u}^{(i)}_{j,n}-u^{(i)}_{j,n})^2 
\label{eq:ks_mse}
\end{equation}
Error growth across rollout horizon is further characterized using time-resolved RMSE.
%============================================
\subsection{Kolmogorov Flow}
\label{subsec:kolmogorov}
%============================================
%---------------------
\subsubsection{Problem and Dataset}
\label{subsubsec:kf_setup}
%---------------------
The final benchmark is two-dimensional Kolmogorov flow governed by the incompressible Navier–Stokes equations and driven by spatial periodic forcing. At moderate Reynolds numbers, the flow develops vortical structures and exhibits a complex chaotic behavior. The vortices stretch and interact with each other to form a much richer non-linear system compared to the other benchmarks. To generate the data, the vorticity–streamfunction form equations (Eqn.~\ref{eq:kf_eqn}) are solved,
\begin{align}
\frac{\partial \omega}{\partial t} + J(\psi,\omega)
&=
\frac{1}{Re_y}\nabla^2\omega + f_\omega(y), \\
\nabla^2 \psi &= -\omega
\label{eq:kf_eqn}
\end{align}
A sinusoidal forcing term is used (Eqn.~\ref{eq:kf_force}),
\begin{equation}
    f_\omega(y)=\sin(k_f y), \qquad k_f=4
\label{eq:kf_force}
\end{equation}
on a periodic domain $[0,2\pi)\times[0,2\pi)$ discretized on a $64\times64$ grid. The spatial derivatives and nonlinear terms are computed pseudospectrally, and time integration uses ETDRK4 with $Re_y=35$ and step $\Delta t_{\mathrm{int}}=0.01$.
Initial conditions are generated from low-pass filtered random Fourier fields spanning multiple dynamical regimes. After a transient of $t_{\mathrm{transient}}=50$, snapshots are recorded over $t_{\mathrm{record}}=30$ with an effective sampling interval
\[
\Delta t_{\mathrm{model}}=0.75 .
\]
Each trajectory contains $N_t=41$ snapshots. Like the KS equations, the dataset includes $600$ trajectories with a train–validation split of 0.8.
%---------------------
\subsubsection{Latent Convolutional Autoencoder Representation}
\label{subsubsec:kolmogorov_latent}
\begin{figure}[]
\centering
\begin{subfigure}{0.48\linewidth}
\centering
\includegraphics[width=\linewidth]{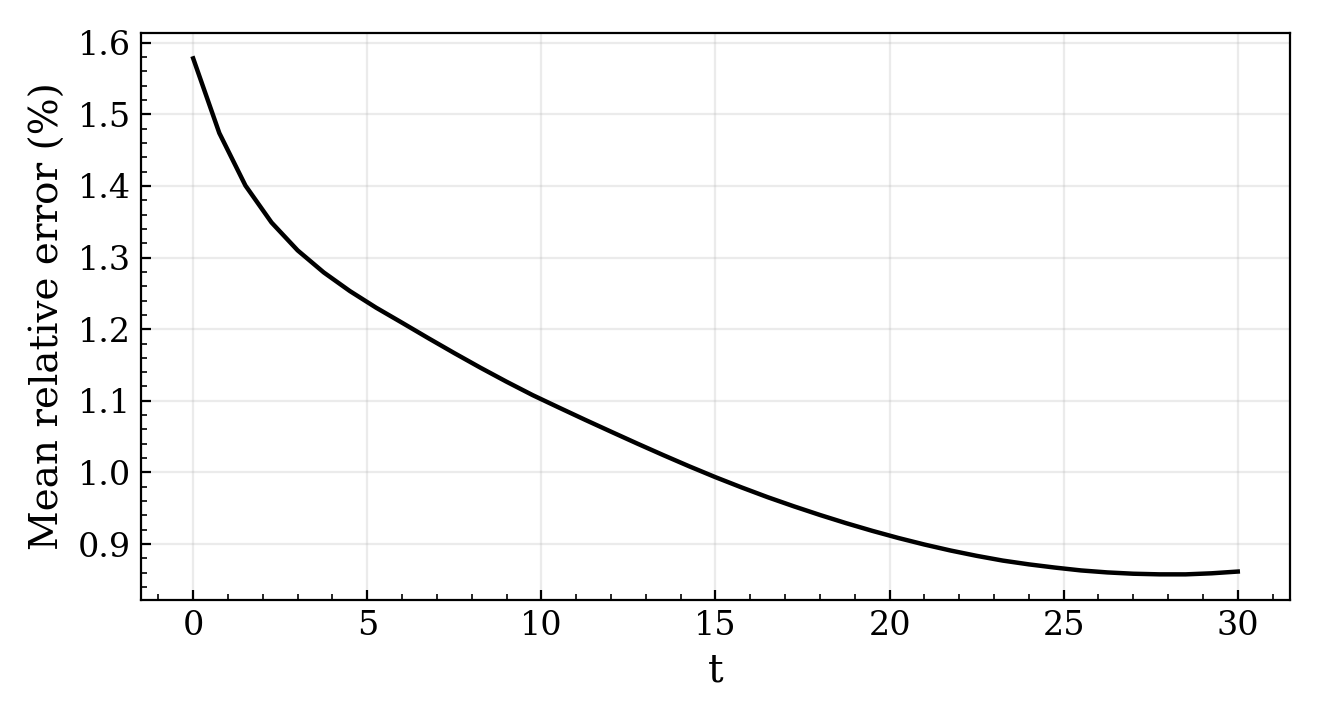}
\end{subfigure}
\hfill
\begin{subfigure}{0.48\linewidth}
\centering
\includegraphics[width=\linewidth]{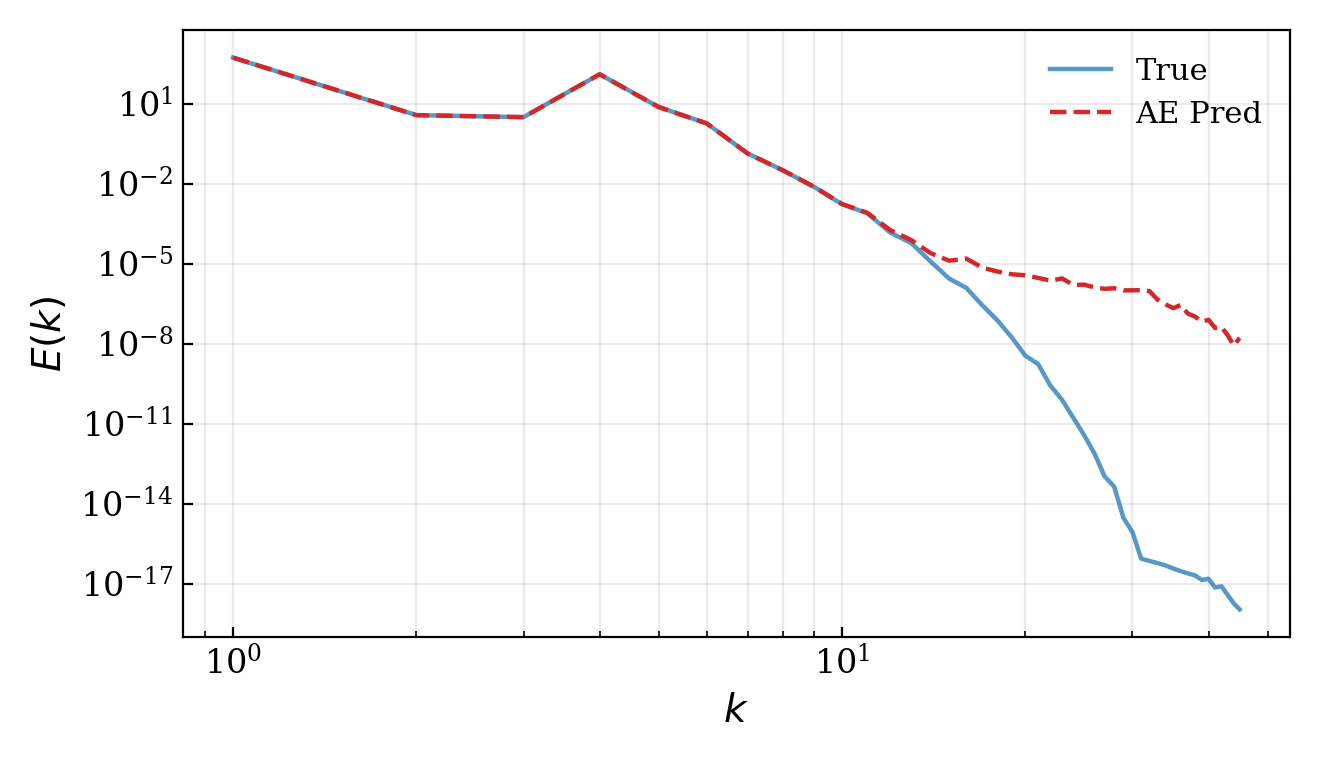}
\end{subfigure}
\caption{\textit{(Left) Mean relative reconstruction error over time on validation trajectories. (Right) Comparison of spatial energy spectra between simulated vorticity fields and autoencoder reconstructions.}}
\label{fig:kf_ae_stats}
\end{figure}

%---------------------
Similar to the KS Equation benchmark, this high-dimensional system is compressed using an autoencoder. However, a convolutional autoencoder is deployed here,
\begin{equation}
\mathbf{z}_t=\mathcal{E}(\tilde{\omega}_t)\in\mathbb{R}^{256},
\qquad
\hat{\omega}_t=\mathcal{D}(\mathbf{z}_t)    
\end{equation}
The encoder consists of three stride-2 convolutional layers followed by a linear projection, and the decoder mirrors this structure with transpose convolutions. The autoencoder strategy is identical to the KS Equations and similar metrics are used to assess the reconstruction quality. Figure~\ref{fig:kf_ae_stats} summarizes the reconstruction quality both in terms of mean relative error and the captured energy.
%============================================
\subsection{Hyperparameter Summary}
\label{subsec:hyperparams}
%============================================
Table~\ref{tab:hyperparams} summarizes the dataset characteristics and training hyperparameters used in this study. For each benchmark, the architectures are configured so that the MLP, LSTM, TCN, NeuralODE, and CoRD have approximately matched parameter counts. As previously highlighted in Section~\ref{subsec:autoreg_lstm}, the LSTM hidden dimension is chosen to produce a comparable parameter budget, accounting for the additional parameters introduced by its gating structure.
\begin{table}[!ht]
\centering
\scriptsize
\begin{tabular}{lccc}
\toprule
\textbf{Parameter} & \textbf{Double Pendulum} & \textbf{KS Eqn.} & \textbf{Kolmogorov Flow} \\
\midrule
\multicolumn{4}{l}{\textit{Dataset}} \\
State dimension
  & $6$
  & $N_x=128 \rightarrow d_z=32$
  & $64\times64 \rightarrow d_z=256$ \\
Training split
  & 0.8
  & 0.8
  & 0.8 \\
Time horizon
  & $10\,\mathrm{s}$
  & $30$ time units
  & $30$ time units \\
\midrule
\multicolumn{4}{l}{\textit{Architecture}} \\
MLP (hidden $\times$ depth)
  & $512 \times 4$
  & $512 \times 6$
  & $512 \times 2$ \\
LSTM (hidden $\times$ layers)
  & $137 \times 4$
  & $512 \times 2$
  & $100 \times 2$ \\
TCN (channels $\times$ blocks)
  & $241 \times 4$
  & $64 \times 2$
  & $110 \times 2$ \\
Neural ODE (hidden $\times$ depth)
  & $512 \times 4$
  & $512 \times 6$
  & $512 \times 2$ \\
CoRD (hidden $\times$ depth)
  & $512 \times 4$
  & $512 \times 6$
  & $512 \times 2$ \\
\midrule
\multicolumn{4}{l}{\textit{Training}} \\
Optimizer
  & \multicolumn{3}{c}{Adam ($\mathrm{lr}=10^{-3}$)} \\
Weight decay
  & \multicolumn{3}{c}{$10^{-4}$} \\
Epochs
  & \multicolumn{3}{c}{1000} \\
Gradient clipping (max norm)
  & \multicolumn{3}{c}{1.0} \\
\bottomrule
\end{tabular}
\caption{\textit{Dataset statistics and hyperparameters for the capacity-matched temporal models across all benchmarks.}}
\label{tab:hyperparams}
\end{table}

%##################################################################################
\section{Results}
\label{sec:results}
%##################################################################################
This section reports the closed loop predictive performances of the neural architectures considered for the study. Two experimental configurations are considered. The primary one is in which the parameter counts are held approximately equal to isolate the architectural effects. Section~\ref{subsec:dp_results},~\ref{subsec:ks_results}, and \ref{subsec:kf_results} detail the findings. This configuration is also used for the mechanism analysis described in Section~\ref{sec:analysis}. In addition, a configuration where each architecture is individually optimized is studied. This approach allows each model to best learn the chaotic systems and is elaborated in Section~\ref{subsec:indiv_models}.
%============================================
\subsection{Double Pendulum}
\label{subsec:dp_results}
%============================================
As discussed in Section~\ref{subsec:dp}, the planar double pendulum is a clean test for rollout stability owing to its strong sensitivity to the initial conditions and non-dissipative nature. Evaluation focuses on error growth over time, aggregate errors, and regime-dependent comparisons between models. 
%---------------------
\subsubsection{Error Growth Over Time}
\begin{figure}[]
    \centering
    \includegraphics[width=0.8\linewidth]{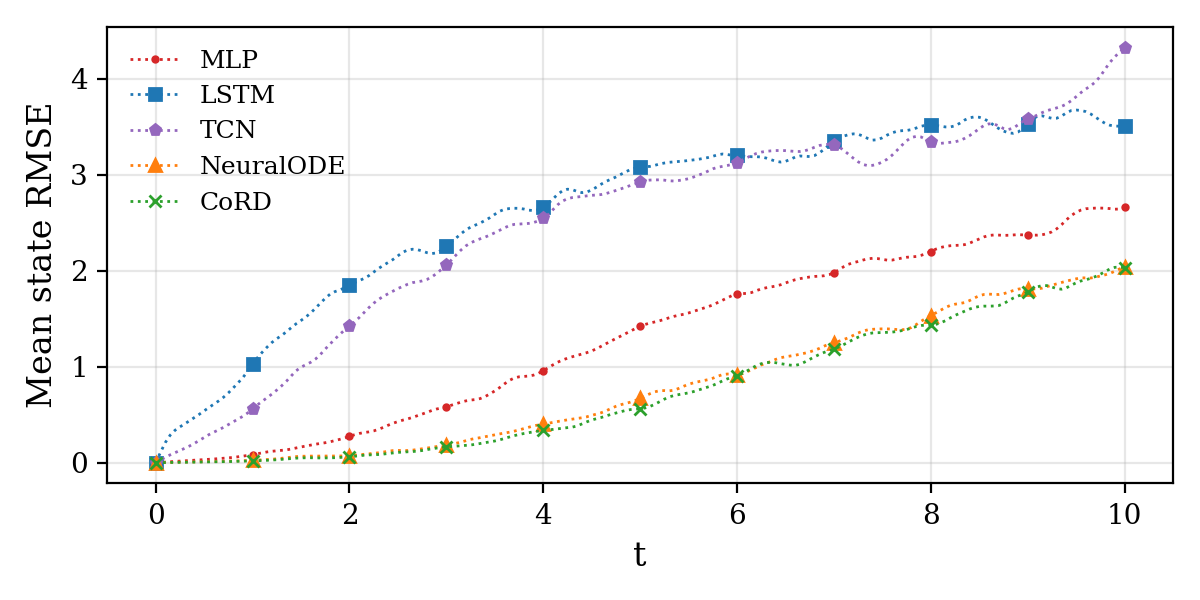}
    \caption{\textit{Double pendulum: Mean state RMSE vs.\ time averaged across validation trajectories.}}
    \label{fig:dp_rmse_vs_time}
\end{figure}

%---------------------
Figure~\ref{fig:dp_rmse_vs_time} shows the mean state RMSE over time averaged across validation trajectories. LSTM and TCN accumulate error most rapidly, exceeding RMSE values of 3 by the end of the 10\,s horizon. The MLP is more stable but still exhibits steady drift. In contrast, the continuous-time models remain substantially more stable, with much slower error growth throughout the rollout. Overall, Neural ODE and CoRD behave quite similarly.
%---------------------
\subsubsection{Aggregate Error}
%---------------------
Table~\ref{tab:dp_metrics} summarizes the aggregated error statistics via MSE for the full and early horizon. NeuralODE achieves the lowest mean full-horizon and early horizon MSE, with CoRD following very closely. Both clearly show stronger predictive capabilities than the discrete-time models. LSTM and TCN exhibit elevated early-horizon errors, consistent with the rapid growth seen in Figure~\ref{fig:dp_rmse_vs_time}.
\begin{table}[!ht]
    \centering
    \scriptsize
    \begin{tabular}{llccc}
        \toprule
        Metric & Model & Mean & Median & P90 \\
        \midrule
        \multirow{5}{*}{Full-horizon state MSE}
            & MLP  & 4.149 & 3.230 & 9.327 \\
            & LSTM & 7.146 & 6.691 & 14.315 \\
            & TCN  & 6.330 & 5.840 & 11.757 \\
            & NeuralODE & 2.109 & 0.821 & 6.128 \\
            & CoRD & 2.276 & 0.831 & 6.617 \\
        \midrule
        \multirow{5}{*}{Early-horizon state MSE (0--2\,s)}
            & MLP  & 0.300 & 0.026 & 0.519 \\
            & LSTM & 2.061 & 0.670 & 6.099 \\
            & TCN  & 2.297 & 1.300 & 5.940 \\
            & NeuralODE & 0.060 & 0.001 & 0.018 \\
            & CoRD & 0.062 & 0.001 & 0.015 \\
        \bottomrule
    \end{tabular}
    \caption{\textit{Double pendulum: trajectory-wise error statistics across the validation trajectories.}}
    \label{tab:dp_metrics}
\end{table}
%---------------------
\subsubsection{Regime-dependent Performance}
\begin{figure}[]
    \centering
    \includegraphics[width=\linewidth]{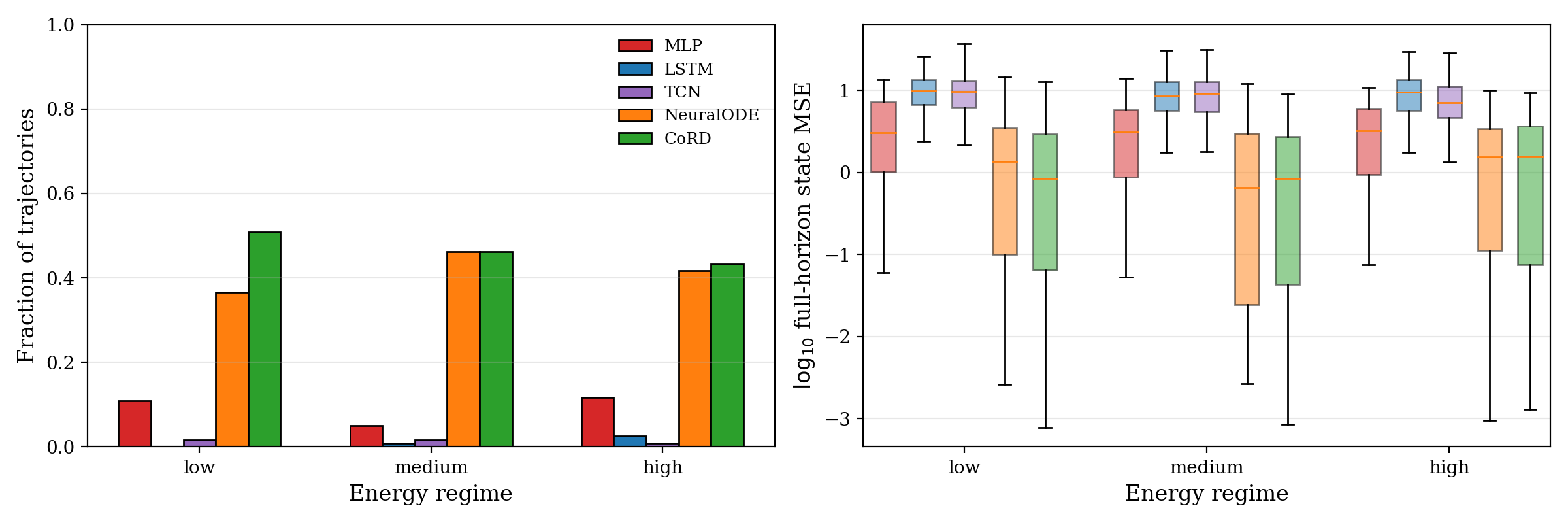}
    \caption{\textit{Double Pendulum Energy-binned regime analysis.
    Left: fraction of validation trajectories on which each model attains the lowest full-horizon MSE within each $E(0)$ tercile. Right: distributions of $\log_{10}$ full-horizon state MSE for each model and regime.}}
    \label{fig:dp_lyap_bins}
\end{figure}

%---------------------
Chaotic systems are dependent on the initial state; therefore, the predictive performance can vary significantly depending on the dynamic regime.  To account for this, validation trajectories are grouped by initial energy $E(0)$ into low, medium, and high energy terciles (Figure~\ref{fig:dp_lyap_bins}). The continuous-time models dominate across all three regimes. CoRD attains the highest win fraction in the low-energy group, with NeuralODE matching performance in the two other categories. The box plots show that both maintain substantially lower median $\log_{10}$ full-horizon MSE than the discrete-time models across all regimes, with CoRD showing slightly tighter dispersion in the medium-energy group.
%============================================
\subsection{KS Equation}
\label{subsec:ks_results}
%============================================
Identical metrics are evaluated for this problem. In this setting, small temporal prediction errors can spread across spatial modes and amplify over time, making long-horizon stability a more demanding test.
%--------------------------------------------
\subsubsection{Qualitative Trajectory Behavior}
\begin{figure}[]
\centering
\includegraphics[width=0.6\linewidth]{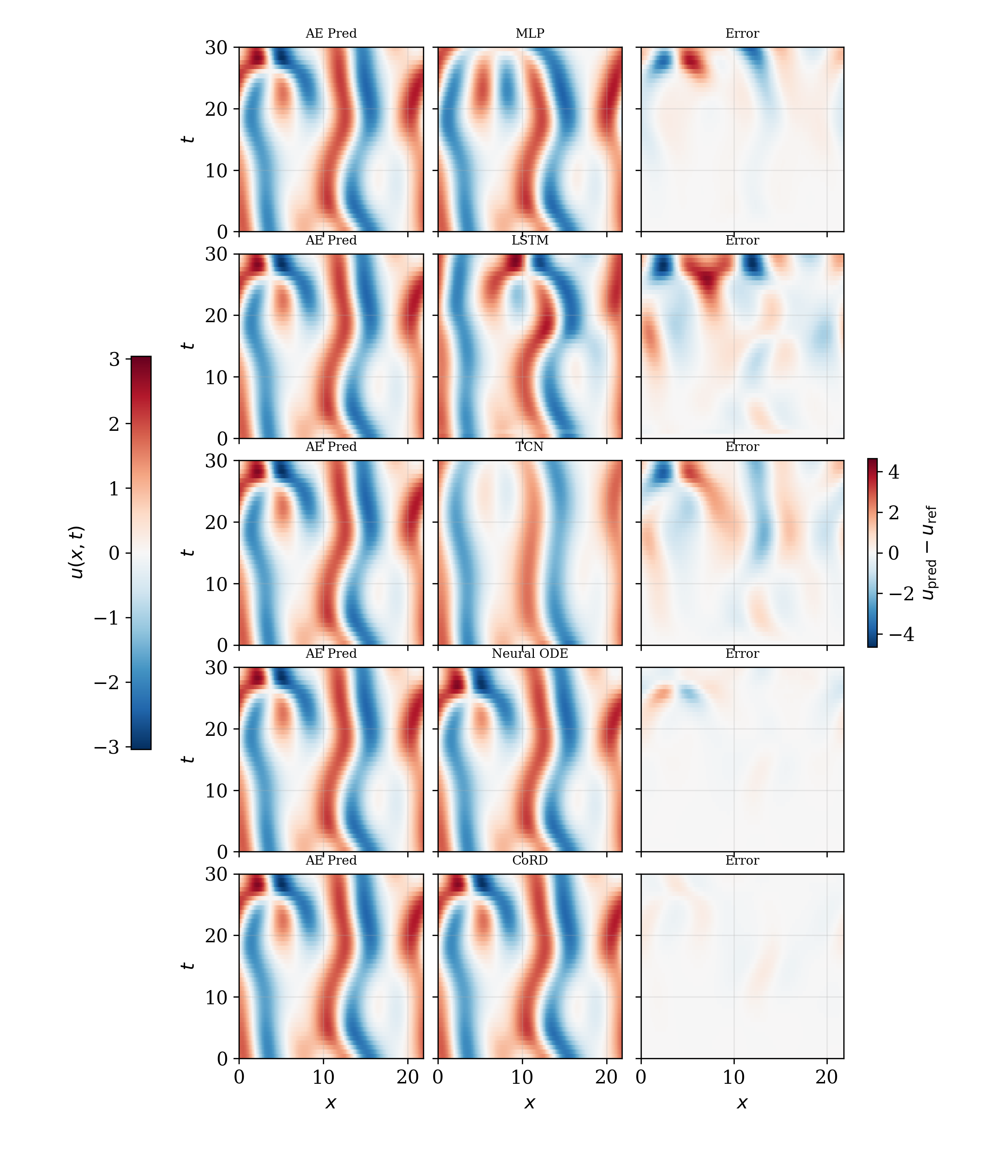}
\caption{\textit{KS Eqn.: Space-time prediction contours for a representative KS validation trajectory.}}
\label{fig:ks_spacetime}
\end{figure}

%--------------------------------------------
Figure~\ref{fig:ks_spacetime} shows the predicted space-time evolution for a randomly chosen validation trajectory. The continuous-time models capture the dominant traveling-wave structures and preserve phase alignment over the full rollout horizon. Their error fields remain largely localized near sharp gradients. However, the discrete-time models progressively lose phase alignment. The MLP shows moderate drift, the LSTM exhibits larger coherent displacement, and the TCN produces the strongest distortion, with both phase and amplitude errors appearing earlier in the rollout.
%--------------------------------------------
\subsubsection{Aggregate Error}
\begin{figure}[]
\centering
\includegraphics[width=0.7\linewidth]{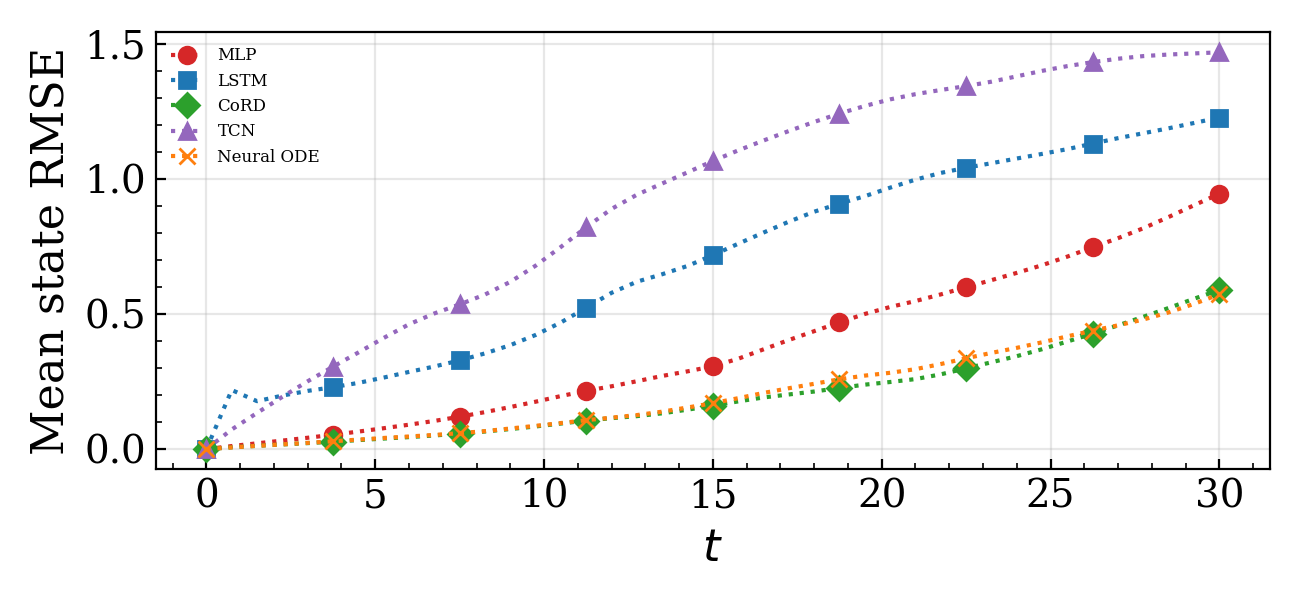}
\caption{\textit{KS Eqn.: Mean state RMSE vs.\ time averaged over validation trajectories.}}
\label{fig:ks_rmse_vs_time}
\end{figure}

%--------------------------------------------
Figure~\ref{fig:ks_rmse_vs_time} shows the mean state RMSE over the rollout horizon averaged across validation trajectories. All models begin with similar short-horizon accuracy, but their trajectories diverge steadily. The TCN exhibits the fastest error growth, followed by the LSTM and MLP. The continuous-time models maintain much slower error accumulation throughout. CoRD is slightly lower at early times, while Neural ODE remains nearly indistinguishable from it over the rest of the horizon. The behavior of the continuous models is consistent with the double-pendulum experiment.
Table~\ref{tab:ks_metrics} summarizes trajectory-wise error statistics. CoRD achieves the lowest median and 90th-percentile full-horizon errors, indicating strong typical and tail performance. Neural ODE yields a slightly lower mean error, reflecting a small number of larger-error outliers for CoRD. Over the early horizon, the differences are much smaller, which indicates that the dominant separation appears during sustained closed-loop rollout rather than in one-step or very short-term prediction.
\begin{table}[!ht]
    \centering
    \scriptsize
    \begin{tabular}{llccc}
        \toprule
        Metric & Model & Mean & Median & P90 \\
        \midrule
        \multirow{5}{*}{Full-horizon state MSE}
            & MLP        & 0.445 & 0.241 & 1.081 \\
            & LSTM       & 0.552 & 0.344 & 1.197 \\
            & TCN        & 0.903 & 0.701 & 1.843 \\
            & NeuralODE  & 0.231 & 0.110 & 0.464 \\
            & CoRD       & 0.194 & 0.095 & 0.408 \\
        \midrule
        \multirow{5}{*}{Early-horizon state MSE}
            & MLP        & 0.054 & 0.052 & 0.076 \\
            & LSTM       & 0.061 & 0.058 & 0.091 \\
            & TCN        & 0.072 & 0.063 & 0.126 \\
            & NeuralODE  & 0.054 & 0.051 & 0.078 \\
            & CoRD       & 0.054 & 0.050 & 0.077 \\
        \bottomrule
    \end{tabular}
    \caption{\textit{KS Eqn.: Trajectory-wise error statistics on the validation set. Mean, median, and 90th percentile are reported for each metric.}}
    \label{tab:ks_metrics}
\end{table}
%--------------------------------------------
\subsubsection{Regime-dependent Performance}
\begin{figure}[]
\centering
\includegraphics[width=0.75\linewidth]{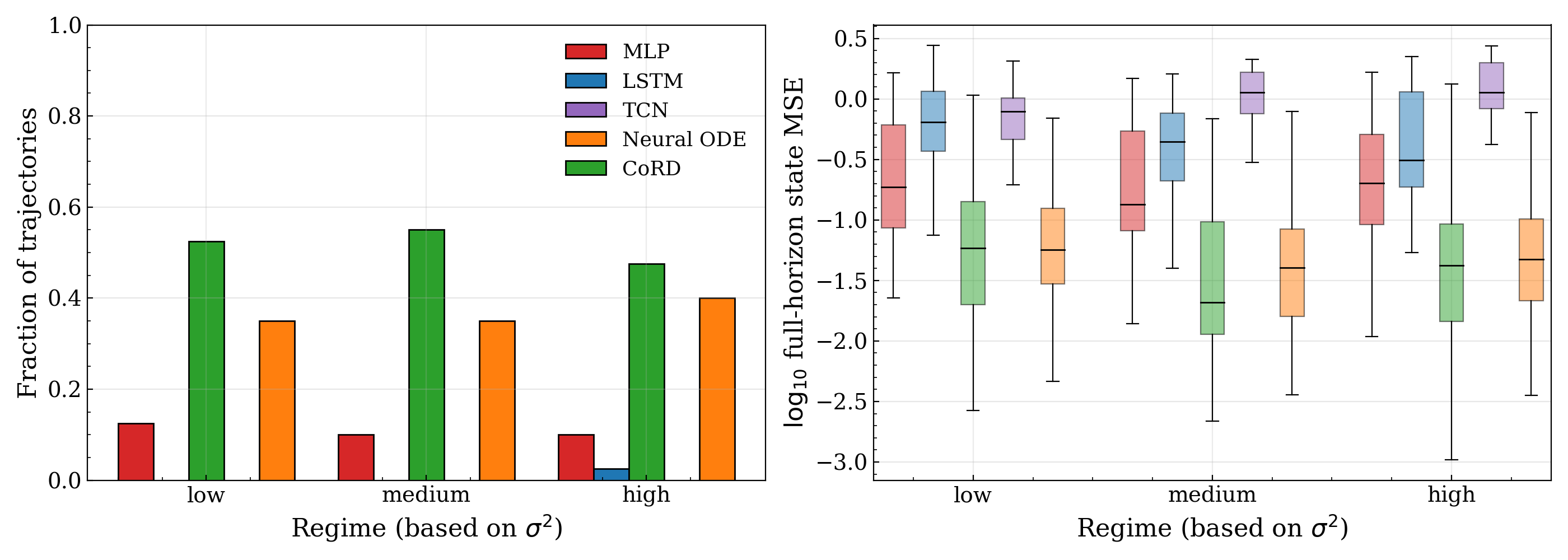}
\caption{\textit{KS Eqn.: Regime analysis using time-averaged spatial variance $\sigma^2$.}}
\label{fig:ks_sigma_bins}
\end{figure}

%--------------------------------------------
Time-averaged spatial variance (Eqn.~\ref{eq:ks_regime_def}) is used to group the trajectories into different regimes,
\begin{equation}
\sigma_i^2 =
\frac{1}{N_t}
\sum_{k=1}^{N_t}
\operatorname{Var}_x\big[u_i(x,t_k)\big]
\label{eq:ks_regime_def}
\end{equation}
Validation trajectories are divided into low, medium, and high variance regimes. Figure~\ref{fig:ks_sigma_bins} shows that the discrete-time models rarely attain the lowest error. CoRD achieves the highest win fraction in the medium variance regime, whereas Neural ODE performs slightly better in the low and high variance groups. In every regime, both continuous-time models maintain lower error distributions than the discrete-time architectures as is observed in the previous experiment.
%============================================
\subsection{Kolmogorov Flow}
\label{subsec:kf_results}
%============================================
Compared to KS Equations, for the Kolmogorov flow, errors are not only amplified but also advected and redistributed spatially, making this benchmark a stringent test of latent temporal stability.
%---------------------
\subsubsection{Trajectory Behavior}
\begin{figure}[]
    \centering
    \includegraphics[width=0.7\linewidth]{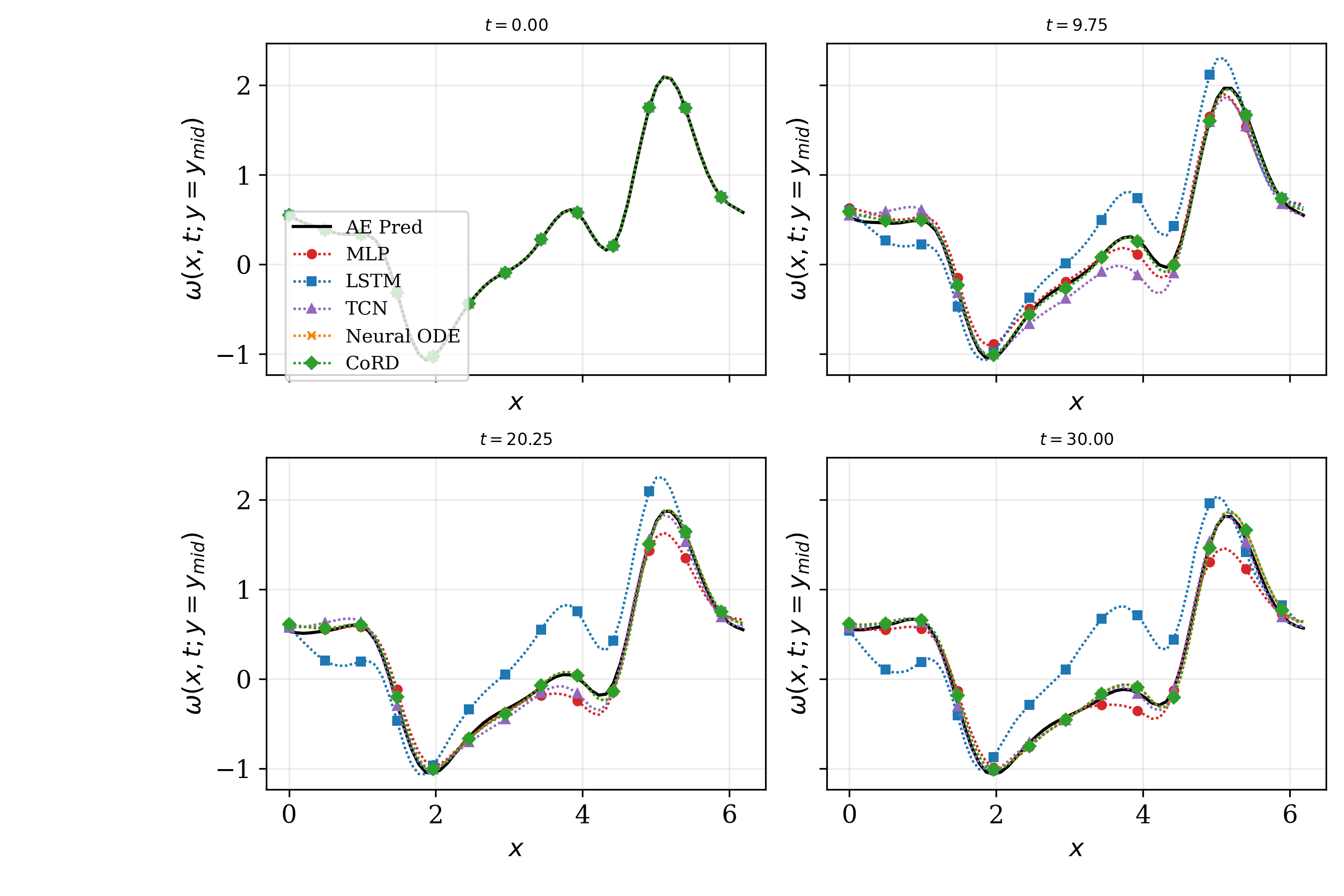}
    \caption{\textit{Kolmogorov flow: Slices of $\omega(x,y_{\mathrm{mid}},t)$ from a representative validation trajectory at four time instants.}}
    \label{fig:kf_lines}
\end{figure}

%---------------------
Figure~\ref{fig:kf_lines} shows one-dimensional slices $\omega(x,y_{\mathrm{mid}},t)$ from a randomly picked validation trajectory. All models coincide with the reference solution at $t=0$ by construction. As the rollout progresses, the discrete-time models begin to accumulate visible phase and amplitude errors. The LSTM exhibits the strongest drift and peak distortion, while the MLP underestimates dominant peaks at later times. The TCN remains more stable than the other discrete-time models but still departs from the reference solution. In contrast, Neural ODE and CoRD remain closely aligned with the reference trajectory throughout the horizon.
%---------------------
\subsubsection{Aggregate Error Statistics}
\begin{figure}[]
    \centering
    \includegraphics[width=0.7\linewidth]{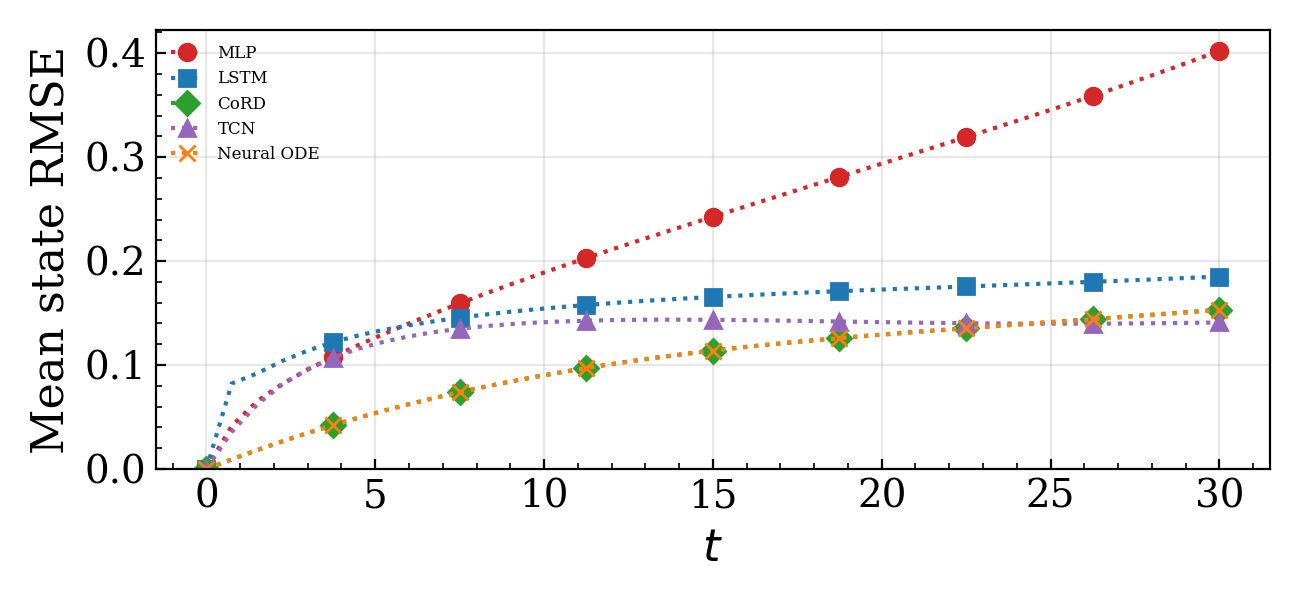}
    \caption{\textit{Kolmogorov flow: Mean state RMSE vs.\ time averaged over validation trajectories.}}
    \label{fig:kf_rmse_vs_time}
\end{figure}

%---------------------
Figure~\ref{fig:kf_rmse_vs_time} shows the mean state RMSE averaged across validation trajectories. The MLP accumulates error steadily and becomes the least accurate model by the end of the rollout. The LSTM exhibits a early spike but settles to a moderate but consistently elevated error level. The TCN error grows more slowly and remains more stable than both MLP and LSTM. Neural ODE and CoRD produce the lowest errors throughout the horizon and are nearly indistinguishable in their RMSE trajectories.
Table~\ref{tab:kf_metrics} confirms this trend quantitatively. Neural ODE and CoRD attain identical mean, median, and upper-tail full-horizon errors, and likewise match on the early-horizon metric. Among the discrete-time models, the LSTM performs worst, while the TCN remains comparatively competitive. Compared to the previous experiments, TCN emerges as a much stronger discrete-time model for capturing the non-linear dynamics. It is likely that the larger receptive field of the TCN proves beneficial for smoother, spatially extended dynamics such as the Kolmogorov flow. For systems that depend more on the instantaneous state, such advantages are not observed.
\begin{table}[!ht]
    \centering
    \scriptsize
    \begin{tabular}{llccc}
        \toprule
        Metric & Model & Mean & Median & P90 \\
        \midrule
        \multirow{5}{*}{Full-horizon state MSE}
            & MLP        & 0.182 & 0.158 & 0.270 \\
            & LSTM       & 0.323 & 0.300 & 0.484 \\
            & TCN        & 0.207 & 0.189 & 0.246 \\
            & NeuralODE  & 0.136 & 0.130 & 0.162 \\
            & CoRD       & 0.136 & 0.130 & 0.162 \\
        \midrule
        \multirow{5}{*}{Early-horizon state MSE}
            & MLP        & 0.179 & 0.156 & 0.280 \\
            & LSTM       & 0.289 & 0.256 & 0.409 \\
            & TCN        & 0.214 & 0.192 & 0.286 \\
            & NeuralODE  & 0.137 & 0.132 & 0.172 \\
            & CoRD       & 0.137 & 0.132 & 0.172 \\
        \bottomrule
    \end{tabular}
    \caption{\textit{Kolmogorov flow: Trajectory-wise state MSE statistics on the validation set.}}
    \label{tab:kf_metrics}
\end{table}
%---------------------
\subsubsection{Regime-dependent Performance}
\begin{figure}[]
    \centering
    \includegraphics[width=0.75\linewidth]{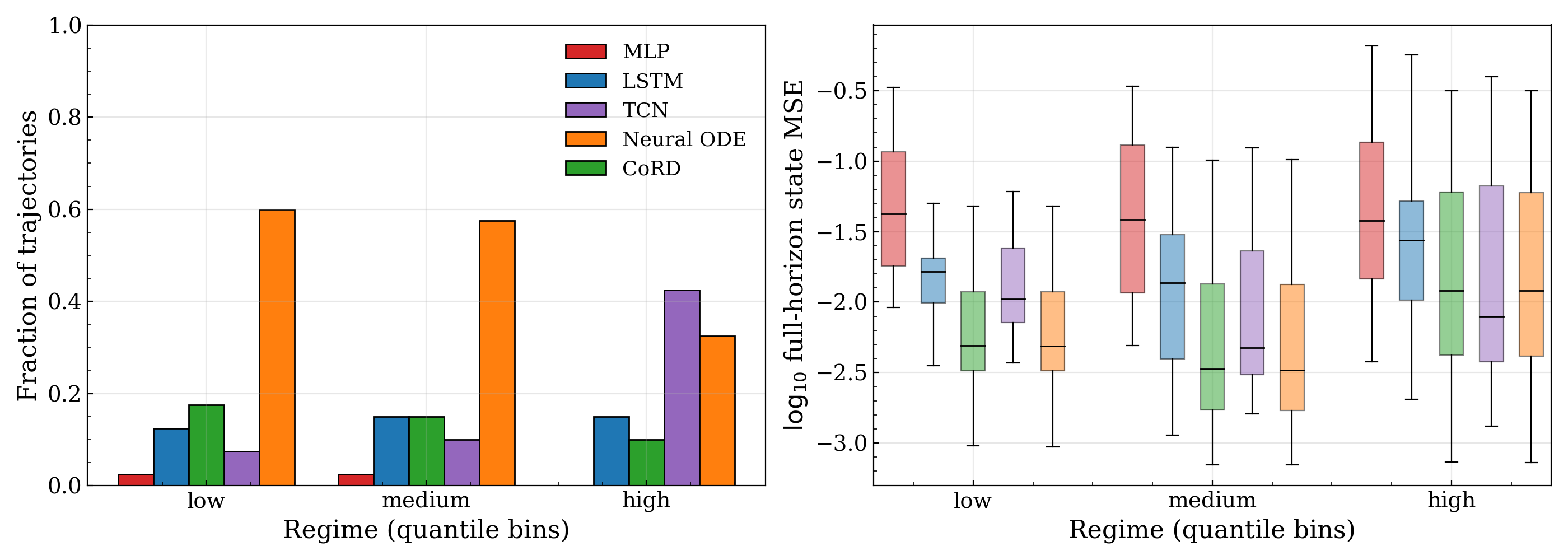}
    \caption{\textit{Kolmogorov flow: regime analysis based on time-averaged enstrophy $\langle \mathcal{E} \rangle$.}}
    \label{fig:kf_enstrophy_bins}
\end{figure}

%---------------------
To examine how performance varies with flow complexity, validation trajectories are grouped by their time-averaged enstrophy,
\begin{equation}
\big\langle \mathcal{E} \big\rangle_i
=
\frac{1}{N_t N_x N_y}
\sum_{n=1}^{N_t} \sum_{j=1}^{N_x} \sum_{k=1}^{N_y}
\tfrac{1}{2}\omega_i^2(x_j,y_k,t_n),
\end{equation}
which measures the intensity of vortical activity. Figure~\ref{fig:kf_enstrophy_bins} shows that the continuous-time models achieve the highest fraction of trajectory-wise wins in the low- and medium-enstrophy regimes. In the high-enstrophy regime, the TCN becomes more competitive and attains the lowest error on a larger fraction of trajectories. Even so, the box plots show that Neural ODE and CoRD retain the lowest median $\log_{10}$ full-horizon MSE across regimes, whereas the MLP and LSTM exhibit higher medians and broader spreads.
%--------------------------------------------
\subsection{Individually Optimized Models}
\label{subsec:indiv_models}
\begin{figure}[]
\centering
\begin{subfigure}{0.32\linewidth}
\centering
\includegraphics[width=\linewidth]{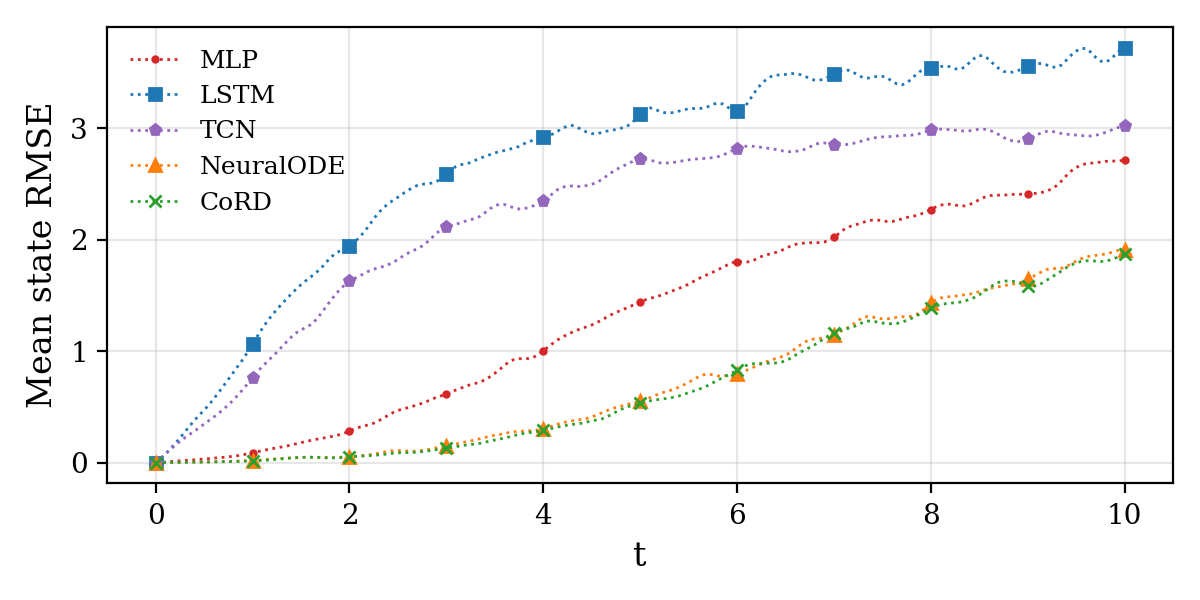}
\caption{Double pendulum}
\label{fig:dp_rmse_indiv}
\end{subfigure}
\hfill
\begin{subfigure}{0.32\linewidth}
\centering
\includegraphics[width=\linewidth]{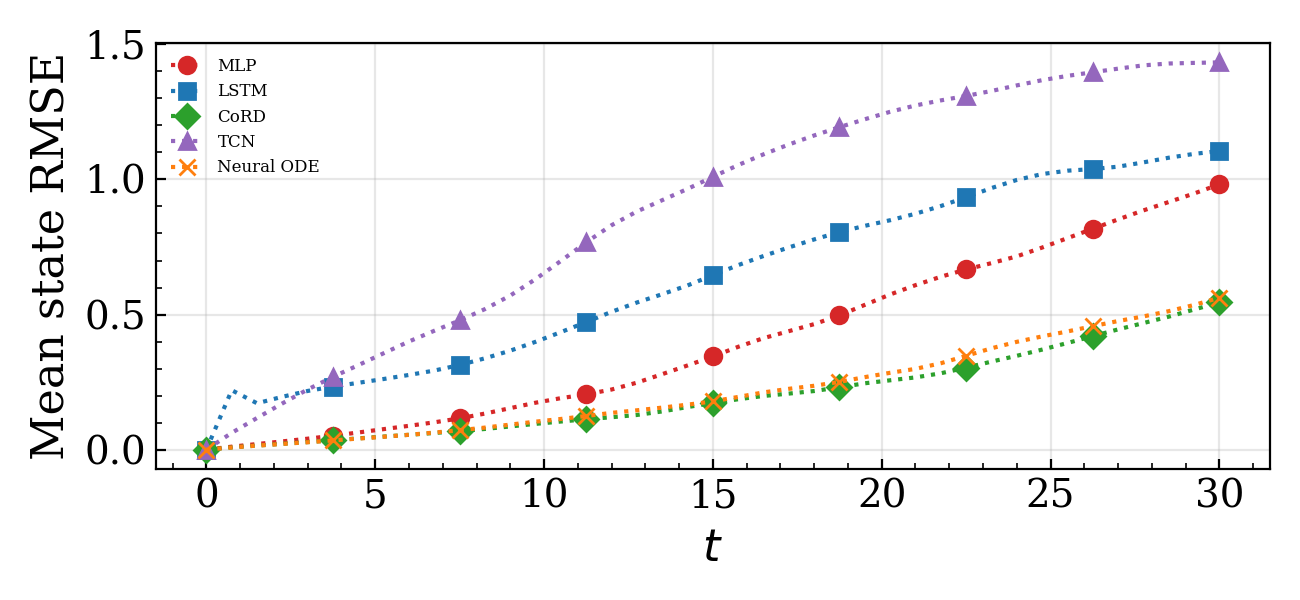}
\caption{KS equation}
\label{fig:ks_rmse_indiv}
\end{subfigure}
\hfill
\begin{subfigure}{0.32\linewidth}
\centering
\includegraphics[width=\linewidth]{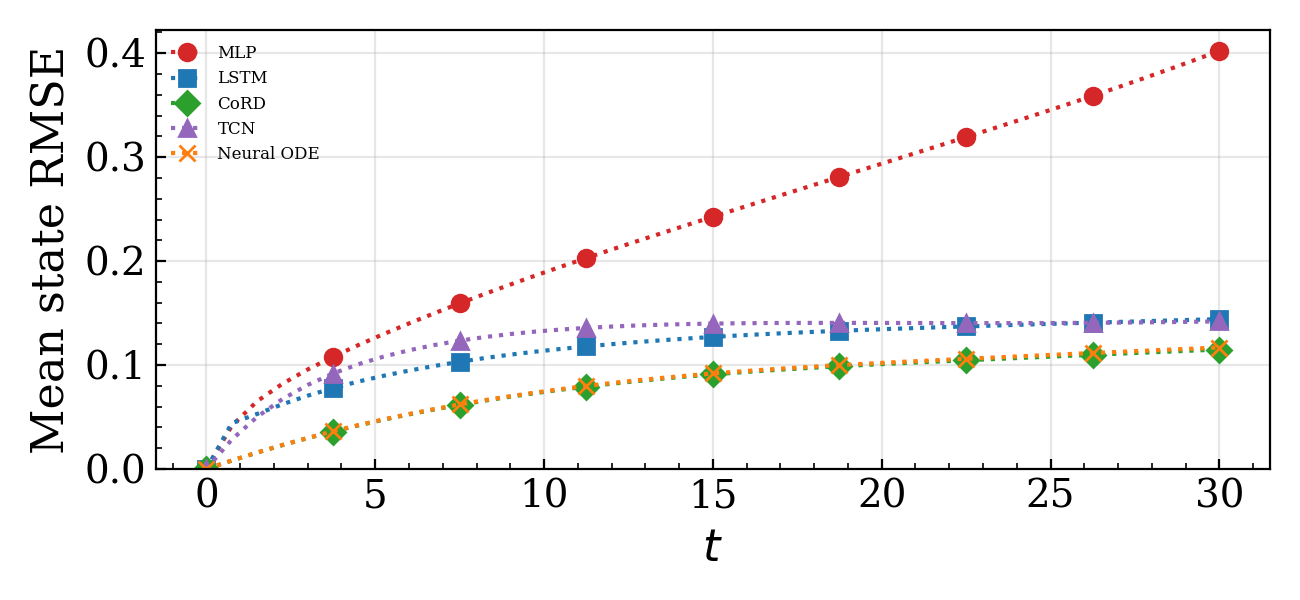}
\caption{Kolmogorov flow}
\label{fig:kf_rmse_indiv}
\end{subfigure}
\caption{\textit{Mean state RMSE vs.\ time for individually optimized models across the three benchmark systems.}}
\label{fig:rmse_indiv_all}
\end{figure}

%--------------------------------------------
The capacity-matched experiments isolate architectural effects by keeping parameter counts comparable across models. To confirm that the observed trends are not driven by this constraint, each architecture is also trained in an individually optimized setting, with hyperparameters tuned for best predictive performance using a pareto-front analysis. Figures~\ref{fig:dp_rmse_indiv}--\ref{fig:kf_rmse_indiv} show the resulting mean state RMSE curves for the three benchmarks.
The overall pattern remains unchanged. Across all the experiments, the continuous-time models display slowest error growth and lowest long-horizon RMSE. The near-indistinguishable behavior of the two continuous-time models also re-emerges. It is therefore safe to conclude that allowing each architecture to operate in its individually optimized configuration does not significantly alter the underlying mechanism with which the models learn.
%##################################################################################
\section{Dynamical Analysis}
\label{sec:analysis}
%##################################################################################
The results presented in the previous section revealed systematic differences in the learning capabilities of the different architectures. However, rollout prediction errors do not fully capture the mechanisms by which the architectures learn. This motivates a closer examination. To attempt that, a few diagnostic tools are leveraged that quantify the introduction and amplification of errors during the closed loop rollout. Sections~\ref{subsec:jac}-\ref{subsec:ft_analysis} focus on local and finite-time error amplification. Specifically, the Jacobian of the learned dynamics, the magnitude of one-step prediction bias, and a finite-time growth metric are evaluated. These quantities describe how perturbations evolve during closed-loop rollout and provide insight into the stability differences observed in the benchmarks. To further characterize the quality of the learned dynamics, it is useful to analyze the attractor geometry learned by the different architecture. Section~\ref{subsec:attaract_struct} provides details on this analysis. KS Equations and Kolmogorov flow are the chosen benchmarks for this study owing to their complex nature.
%---------------------
\subsection{One-Step Jacobian}
\label{subsec:jac}
%---------------------
Let $E(\cdot)$ denote the frozen autoencoder encoder and define the raw latent trajectory
\[
z^{\mathrm{raw}}_{i,t}=E(u_{i,t})\in\mathbb{R}^D .
\]
For each model $m$, evaluation is performed in the model’s training coordinates using its
normalization statistics $(\mu_m,\sigma_m)$,
\begin{equation}
z^{(m)}_{i,t}=\frac{z^{\mathrm{raw}}_{i,t}-\mu_m}{\sigma_m},
\qquad
c^{(m)}_i=z^{(m)}_{i,0}.
\end{equation}
Operating points are sampled as $(i,t)$ with $t\in[r_f-1,\,N_t-2]$, where $r_f$ denotes
the TCN receptive field. The one-step mapping is,
\begin{equation}
z^{(m)}_{i,t+1}=F_{\theta_m}\!\left(z^{(m)}_{i,t};\,\mathcal{C}^{(m)}_{i,t}\right),
\end{equation}
where $\mathcal{C}^{(m)}_{i,t}$ denotes architecture-consistent conditioning constructed
from ground-truth latent trajectories.
The local Jacobian of the learned dynamics is,
\begin{equation}
J^{(m)}_{i,t}=
\frac{\partial}{\partial z^{(m)}_{i,t}}
F_{\theta_m}\!\left(z^{(m)}_{i,t};\,\mathcal{C}^{(m)}_{i,t}\right),
\end{equation}
computed with gradients taken only with respect to $z^{(m)}_{i,t}$.
Local perturbation amplification is summarized by the spectral radius,
\begin{equation}
\rho(J^{(m)}_{i,t})=\max_j |\lambda_j(J^{(m)}_{i,t})|,
\end{equation}
where $\rho>1$ indicates local expansion and $\rho<1$ contraction.

\begin{figure}[]
\centering
\includegraphics[width=0.48\linewidth]{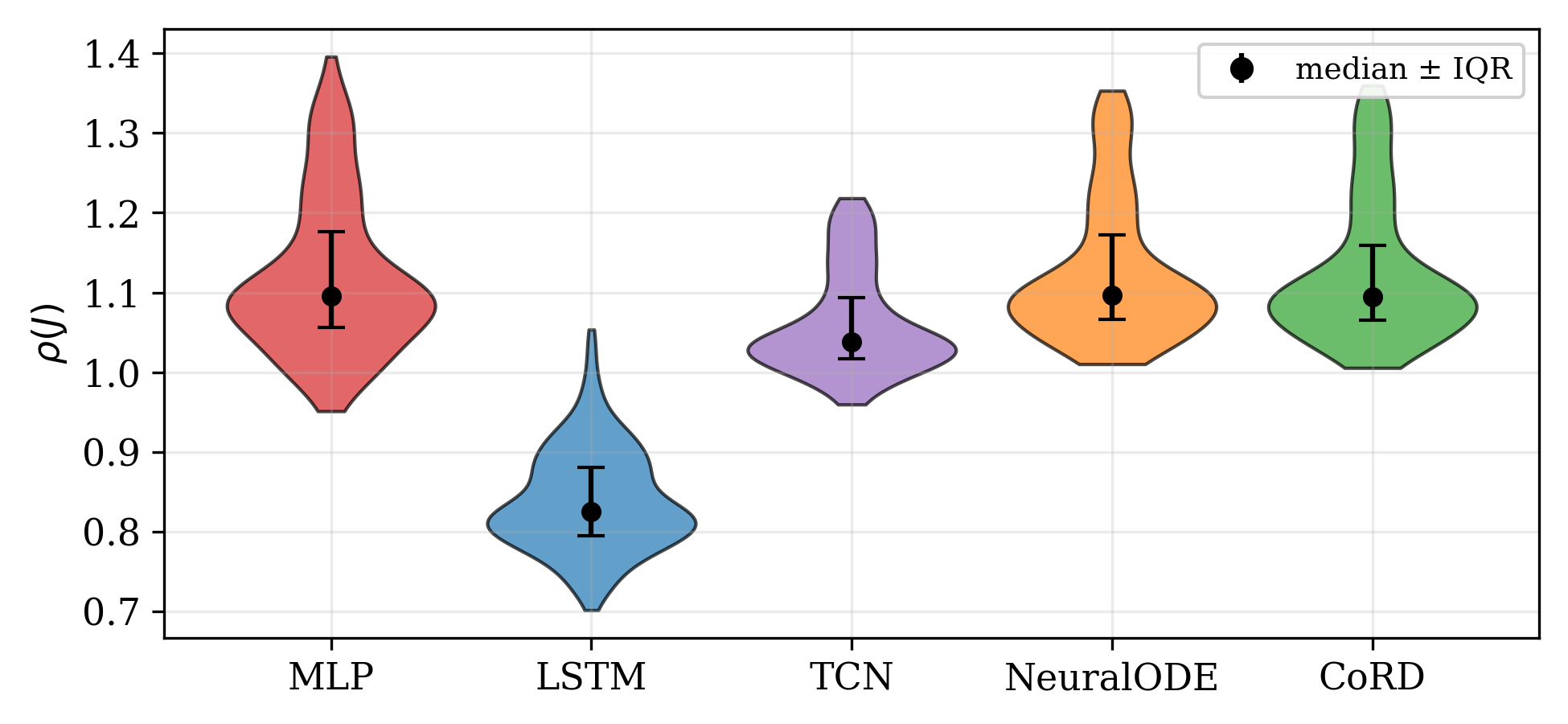}
\includegraphics[width=0.48\linewidth]{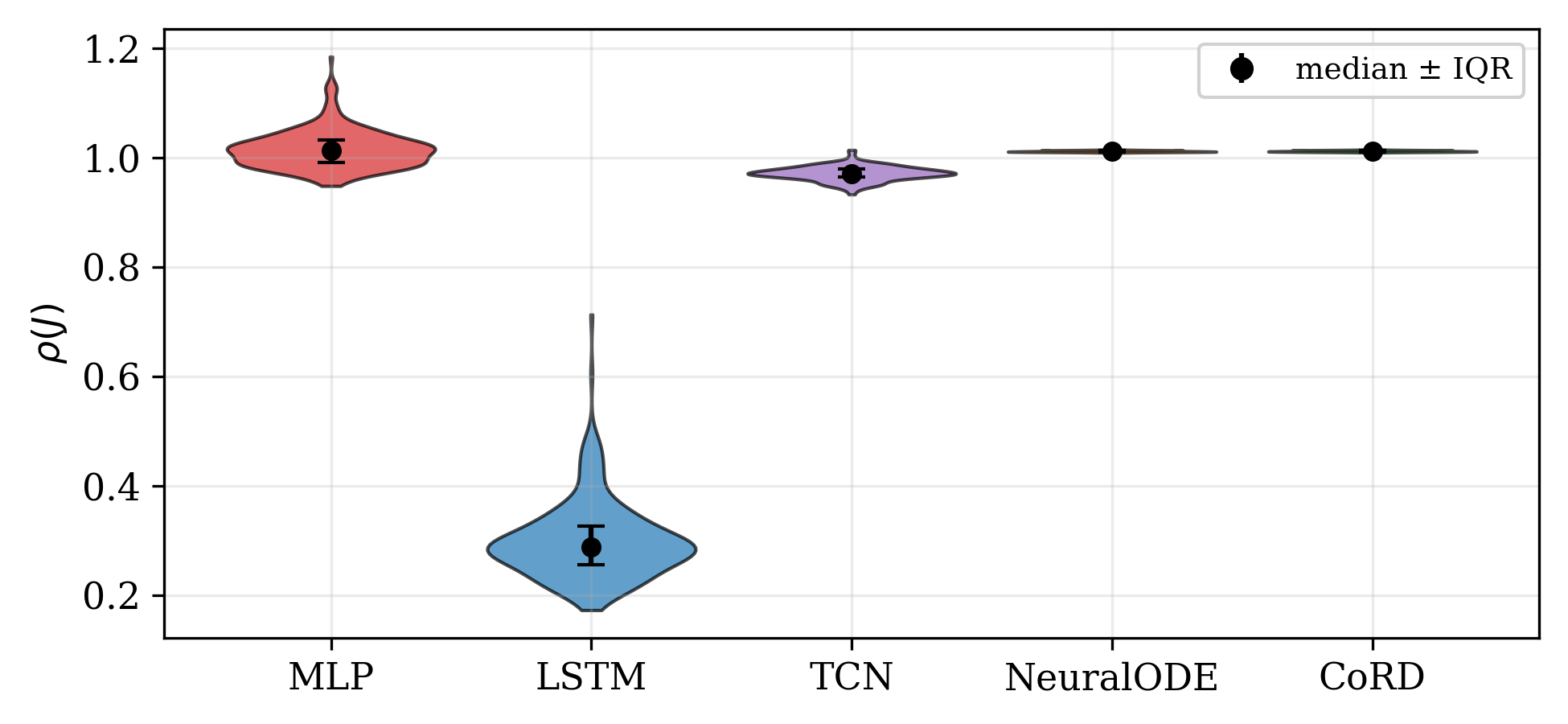}
\caption{\textit{Distributions of the Jacobian spectral radius $\rho(J)$ for the KS equation
(left) and Kolmogorov flow (right). Points denote median values and error bars indicate
the interquartile range.}}
\label{fig:jacobian_pde}
\end{figure}

Figure~\ref{fig:jacobian_pde} shows the distributions of $\rho(J)$. It is evident that the models are sensitive to local perturbations. LSTM shows strongly contracted Jacobians with median spectral radii below 1. This indicates that the architecture has a tendency to damp perturbations. While this might present an appearance of stability, this could lead to erroneous predictions as chaotic systems often require controlled growth of perturbations. This means that the perturbations should grow at a rate consistent with the system's positive Lyapunov exponent. Pure damping can lead a model to learn a dissipative approximation of the chaotic dynamics. MLP has $\rho(J)>1$, indicating that small deviations can possibly grow at each step. However, it is important to note that unchecked growth can force the trajectories to deviate significantly from the attractor and result in unbounded predictions. The TCN remains close to marginal stability, with spectral radii clustered near unity. Continuous time models display similar properties with some differences across the two benchmarks. For KS Equations, the violin plots hint that the perturbations may either grow or decay depending on the local region of the attractor. For Kolmogorov flow, the distribution is tighter around 1, implying that perturbations are neither strongly amplified or damped. These results summarize the effect of local sensitivity across the different architectures. These do not fully explain the rollout behavior as long-horizon error also depends on systematic prediction bias and cumulative effect of repeated model updates. These will be examined in the next sections. 
%---------------------
\subsection{Relative Error Injection}
\label{subsec:rel_error}
%---------------------
Relative one-step bias measures the magnitude of prediction error introduced at each application of the learned map. For sampled operating points $(i,t)$ constructed from ground-truth latent trajectories, the bias metric is defined as,
\begin{equation}
B(\theta_m)
=
\mathbb{E}_{i,t}
\left[
\frac{
\| \hat z^{\mathrm{raw}}_{i,t+1}-z^{\mathrm{raw}}_{i,t+1} \|_2
}{
\| z^{\mathrm{raw}}_{i,t+1} \|_2 + \varepsilon
}
\right],
\end{equation}
where predictions are produced in normalized coordinates and subsequently mapped back to raw latent space before computing the error. Results are reported as $\log_{10} B(\theta_m)$.

\begin{figure}[]
\centering
\begin{subfigure}{0.48\linewidth}
\centering
\includegraphics[width=\linewidth]{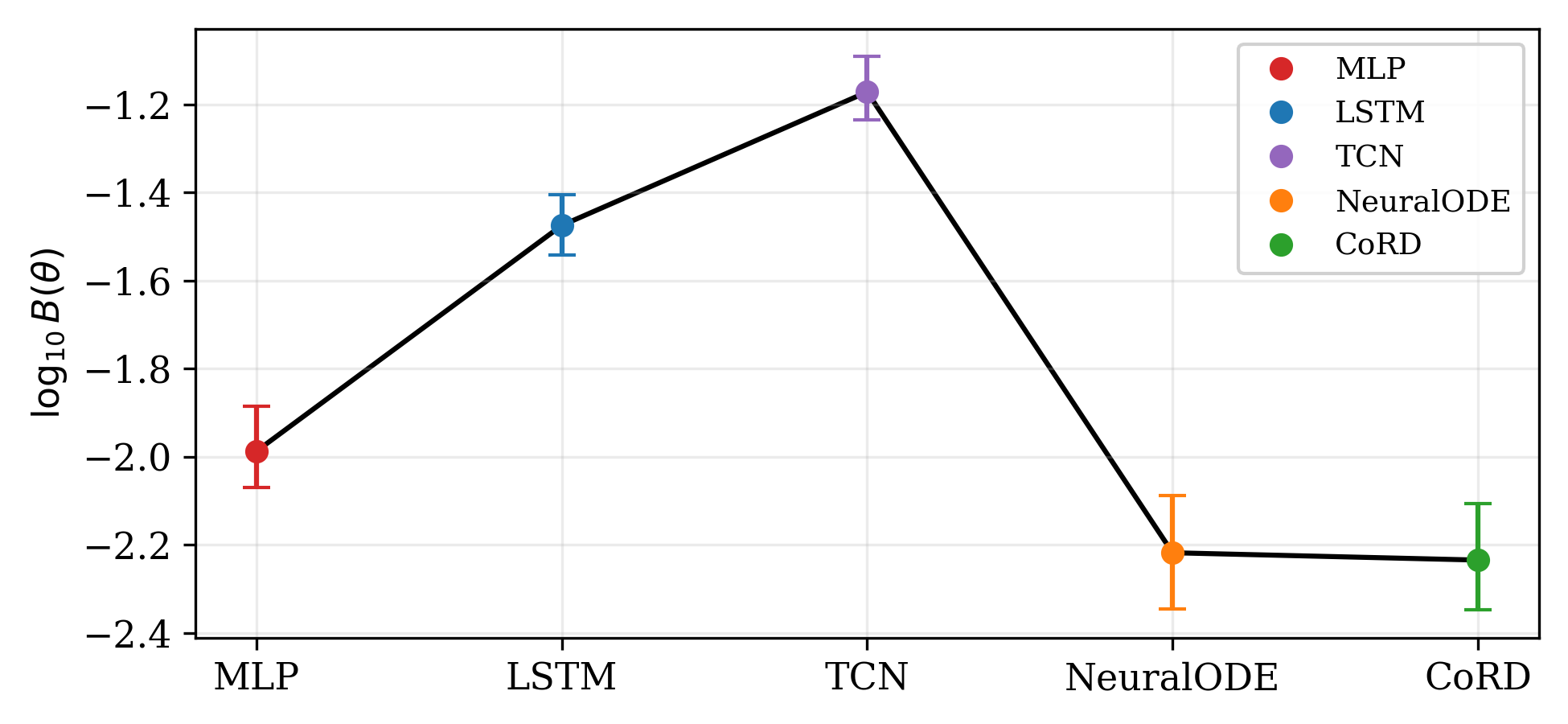}
\caption{KS equation}
\label{fig:ks_bias}
\end{subfigure}
\hfill
\begin{subfigure}{0.48\linewidth}
\centering
\includegraphics[width=\linewidth]{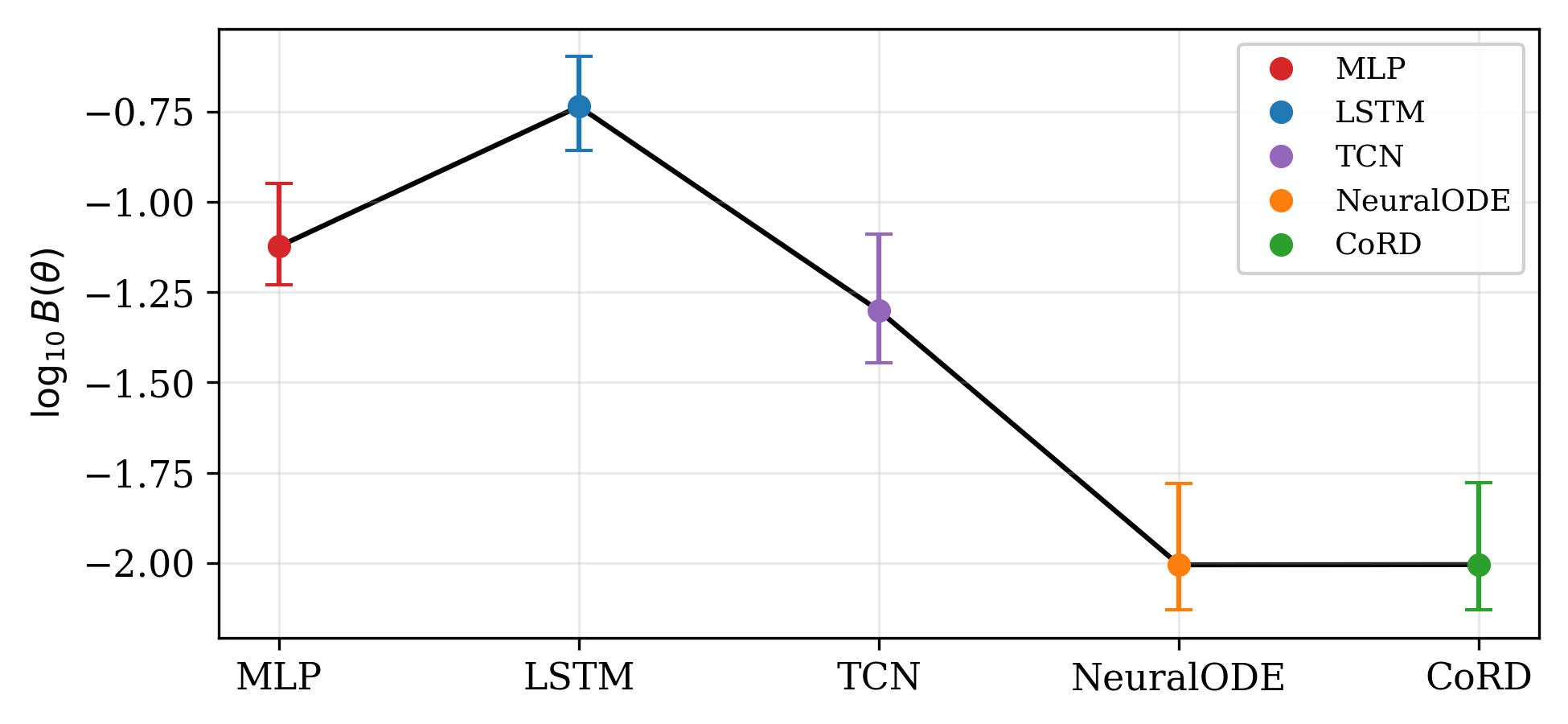}
\caption{Kolmogorov flow}
\label{fig:kf_bias}
\end{subfigure}
\caption{\textit{Relative one-step bias $\log_{10} B(\theta)$ for the latent PDE models.}}
\label{fig:bias_pde}
\end{figure}

Figures~\ref{fig:ks_bias} and~\ref{fig:kf_bias} show the one-step bias for both systems. Continuous-time models operate with the lowest bias. This is likely due to the neural architecture. Instead of learning a discrete jump from one step to the other, this class of models integrate a learned vector field, which in turn keeps each step update closely anchored to the underlying dynamics. MLP falls somewhat in the middle. Despite learning a discrete jump, it lacks the complexity of sequence modeling which could limit the bias at each step. LSTM and TCN models operate in a much different scenario with the highest biases. LSTMs carry long-range memory through gating mechanisms. Their internal state is optimized for temporal coherence over longer windows rather than precise instantaneous accuracies. TCNs have a similar issue with the dilated convolutions aggregating information over time. This can smooth over the fine grained local dynamics that are particularly important for one-step update accuracies. The outcome is both models perform step updates with non-negligible errors that continue to get compounded. Combining with the Jacobian analysis, this section explains the learning mechanisms of the models more clearly. A model with high bias but a contractive Jacobian will drift slowly but steadily toward a degenerate attractor. A model with lower bias but an expansive Jacobian may track well initially before amplifying small errors into divergence. Understanding the combination of these properties, how much error is introduced at each step and how aggressively that error is amplified, is what actually governs long-horizon behavior. This is examined next.
%---------------------
\subsection{Finite-Time Model Lyapunov Exponent}
\label{subsec:ft_analysis}
%---------------------
Nonlinear multi-step amplification is evaluated using a closed-loop finite-time Lyapunov exponent (FTLE). The spectral radius analysis in section~\ref{subsec:jac} measured how infinitesimally small perturbations behave at a single point under one model step. This metric quantifies how finite perturbations grow over many steps along an actual trajectory. This analysis is important as chaotic systems are defined by the global sensitivity, i.e., a perturbation that starts small, might drift into a state space region with completely different local dynamics. FTLE is an appropriate tool to assess this amplification.
For an operating point $(i,t)$, a perturbation of magnitude $\varepsilon$ is applied in raw
latent space,
\begin{equation}
z^{\mathrm{raw,pert}}_{i,t}
=
z^{\mathrm{raw}}_{i,t}+\delta z,
\qquad
\|\delta z\|_2=\varepsilon 
\end{equation}
Both the reference and perturbed states are propagated for $K$ closed-loop steps. This is a stricter test because any error the model introduces is immediately embedded in the next input state. The resulting separation is measured in raw latent space,
\begin{equation}
\Lambda_K^{(10)}(\theta_m)
=
\frac{1}{K}
\log_{10}
\left(
\frac{
\| z^{\mathrm{raw,pert}}_{i,t+K}-z^{\mathrm{raw}}_{i,t+K} \|_2
}{
\varepsilon
}
\right).
\end{equation}
A value of $\Lambda_K^{(10)} = 0$ means the perturbation neither grew nor shrank over the rollout; a value of 1 indicates a tenfold amplification per step on average; negative values indicate that the model is contracting perturbations.
\begin{figure}[]
\centering
\begin{subfigure}{0.48\linewidth}
\centering
\includegraphics[width=\linewidth]{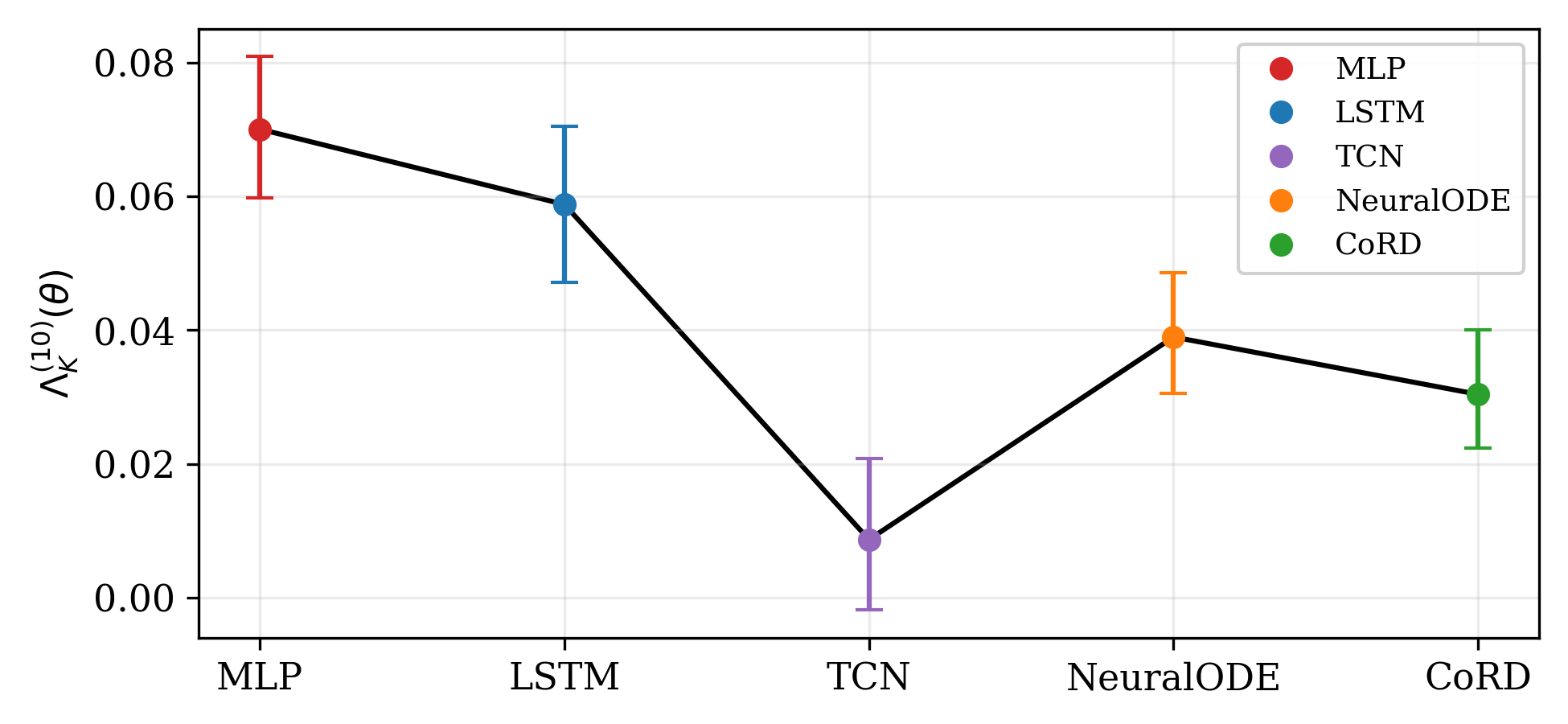}
\caption{KS equation}
\label{fig:ks_lambda}
\end{subfigure}
\hfill
\begin{subfigure}{0.48\linewidth}
\centering
\includegraphics[width=\linewidth]{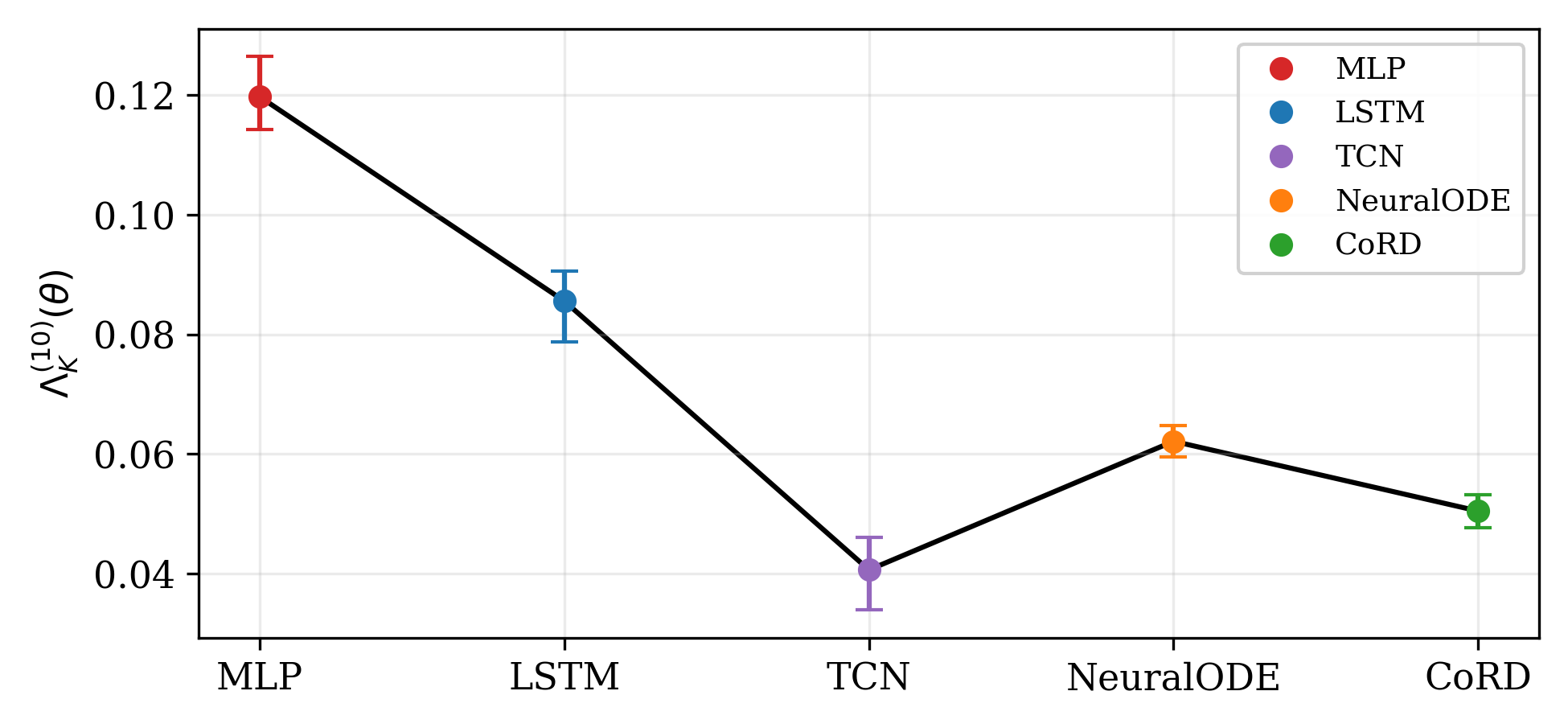}
\caption{Kolmogorov flow}
\label{fig:kf_lambda}
\end{subfigure}
\caption{\textit{Finite-time model Lyapunov exponent $\Lambda_K^{(10)}(\theta)$.}}
\label{fig:lambda_pde}
\end{figure}
Figures~\ref{fig:ks_lambda} and~\ref{fig:kf_lambda} show the resulting growth statistics. Across both PDE systems the continuous-time models exhibit the smallest perturbation growth rates, indicating weaker nonlinear amplification during rollout. This is coherent with their low one-step bias and near-marginal Jacobian behavior: because they track the reference trajectory closely at each step, perturbations are less likely to be steered into high-expansion regions of the attractor. The MLP shows substantially larger growth exponents, consistent with its Jacobian $(\rho > 1)$. This implies that systematic local amplification at each step compounds into large trajectory-level divergence over the rollout horizon. The TCN occupies an intermediate position. The LSTM result deserves particular attention because it appears to conflict with the Jacobian analysis. Recall that the LSTM produced a contractive spectral radius $(\rho < 1)$, suggesting that it damps local perturbations. Yet here it shows among the largest FTLE values. This apparent contradiction is resolved by recognizing that the FTLE operates under closed-loop rollout, which is a different regime than the one-step Jacobian. Under closed-loop rollout, the LSTM's comparatively high one-step bias causes its trajectory to drift away from the reference states. The Jacobian statistics computed on the true trajectory may say very little about the local dynamics at the states the LSTM actually visits during rollout. From those drifted positions, the dynamics may be expansive. Furthermore, the finite-amplitude FTLE perturbation can grow large enough to exit the contractive basin that the linearized Jacobian was measuring, and could be entering regions the infinitesimal analysis never probed. The FTLE integrates all of this, trajectory drift, finite-amplitude nonlinearity, and multi-step compounding, which is precisely what makes it a complementary diagnostic to the Jacobian.

The bias $B(\theta_m)$ and $\Lambda_K^{(10)}(\theta_m)$ describe two distinct but interacting failure modes. Bias determines how far off the attractor each prediction step begins, FTLE determines how aggressively those initial offsets are amplified during rollout. A model that injects large errors at each step and then amplifies them faces the worst of both worlds, while a model that is both accurate per-step and non-amplifying can sustain coherent trajectories over long horizons. This mechanistic picture provides a direct explanation for the RMSE curves reported in Section~\ref{sec:results}. The slower error growth of the continuous-time models is not coincidental but a downstream consequence of their simultaneously low bias and weak perturbation amplification operating together across the rollout.
%---------------------
\subsection{Attractor Structure}
\label{subsec:attaract_struct}
%---------------------
A surrogate that effectively learns the chaotic dynamics should not only be able to predict accurate trajectories over short-horizons but also reproduce the invariant structure of the underlying dynamics. This section presents a quantification of the same. Attractor diagnostics are computed in the decoded physical space.

For both PDE systems, the reference cloud is drawn from the autoencoder reconstructions rather than the raw simulated fields. All models operate on the same latent space and decode through the frozen autoencoder. Therefore, any attractor distortions can be solely attributed to the temporal model dynamics rather than the deficiencies of the autoencoder reconstructions. A PCA basis is then fitted on the reference cloud to identify the directions of maximum variance and to construct an optimal low-dimensional projection for comparison. All model attractors are projected into this common space.

\begin{figure}[]
\centering
\begin{subfigure}{0.4\linewidth}
\centering
\includegraphics[width=\linewidth]{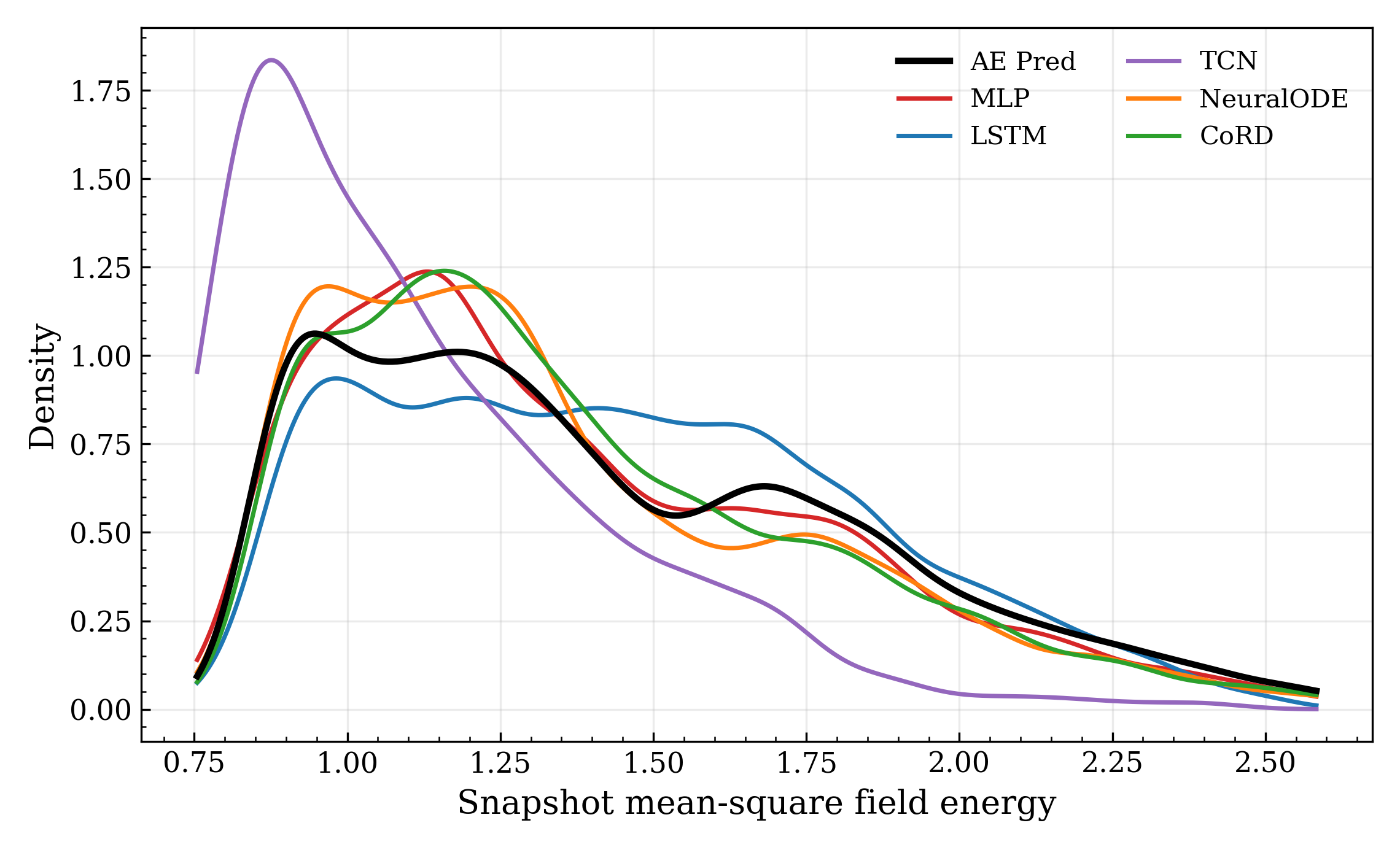}
\end{subfigure}
\vspace{0.5em}
\begin{subfigure}{0.6\linewidth}
\centering
\includegraphics[width=\linewidth]{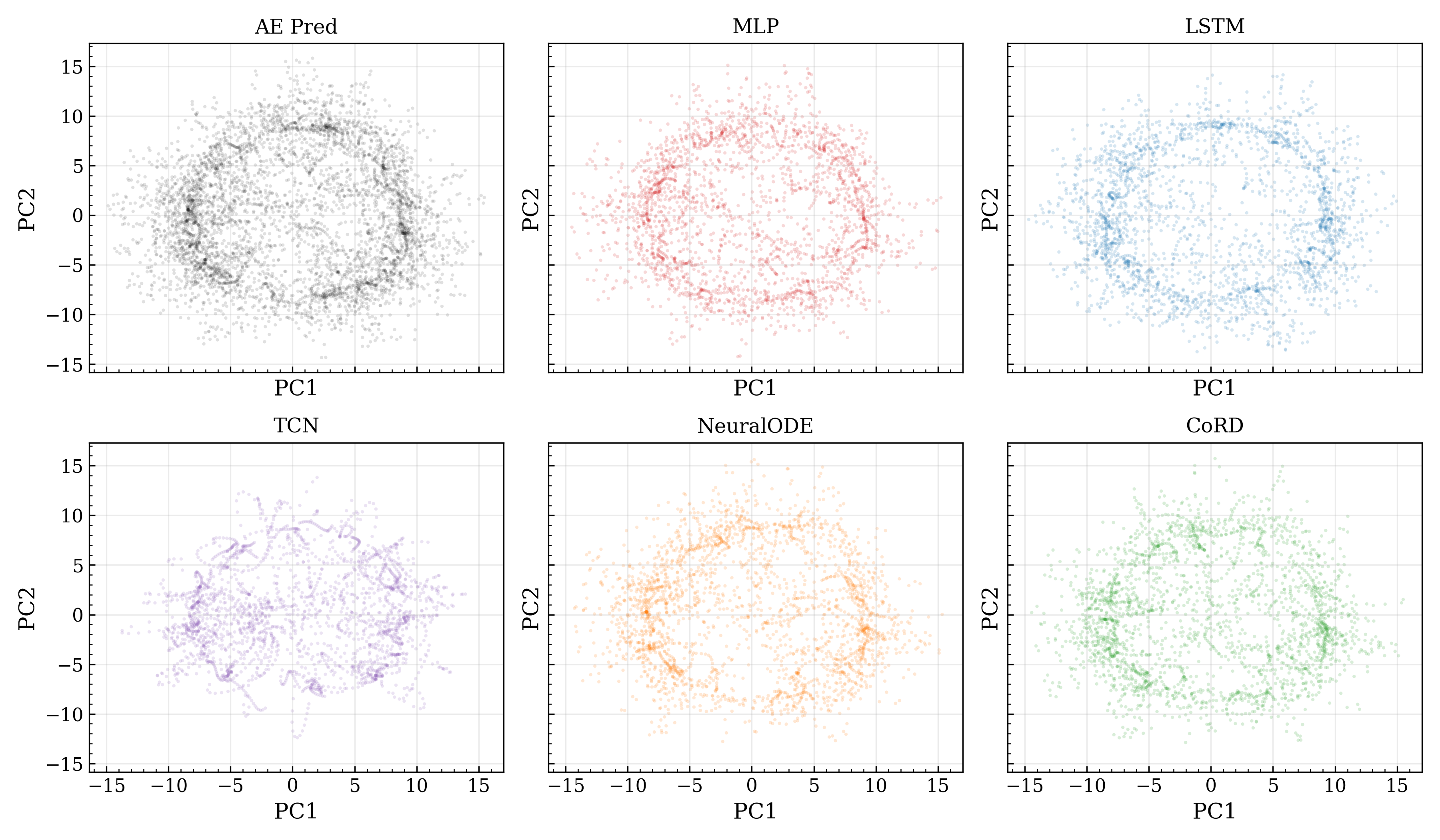}
\end{subfigure}
\caption{\textit{KS attractor diagnostics: Top: kernel density estimates of the snapshot spatial
variance \(\sigma^2\). Bottom: PCA projections of the attractor clouds in a basis fit on the
reference AE reconstruction cloud.}}
\label{fig:ks_attractor}
\end{figure}

For the KS Equation, Figure~\ref{fig:ks_attractor} compares the distribution of snapshot spatial variance \(\sigma^2\) and the PCA projections. Spatial variance is a proxy quantification of the field energy and complexity, where low values correspond to smooth, less chaotic states, while the higher values reflect the energetic, multi-scale states. The continuous-time models and MLP remain closest to the reference distribution that has a ring like shape, accurately representing the energy spread across states. The TCN shows a distorted cloud, indicating that the model learns a much smoother dynamic than what the system actually entails. LSTM preserves the overall distribution shape; however, it has a broader spread. This is reflective of the greater trajectory drifts, i.e., the high one-step bias causes closed-loop rollouts to settle into states outside the true attractor.
\begin{figure}[]
\centering
\begin{subfigure}{0.4\linewidth}
\centering
\includegraphics[width=\linewidth]{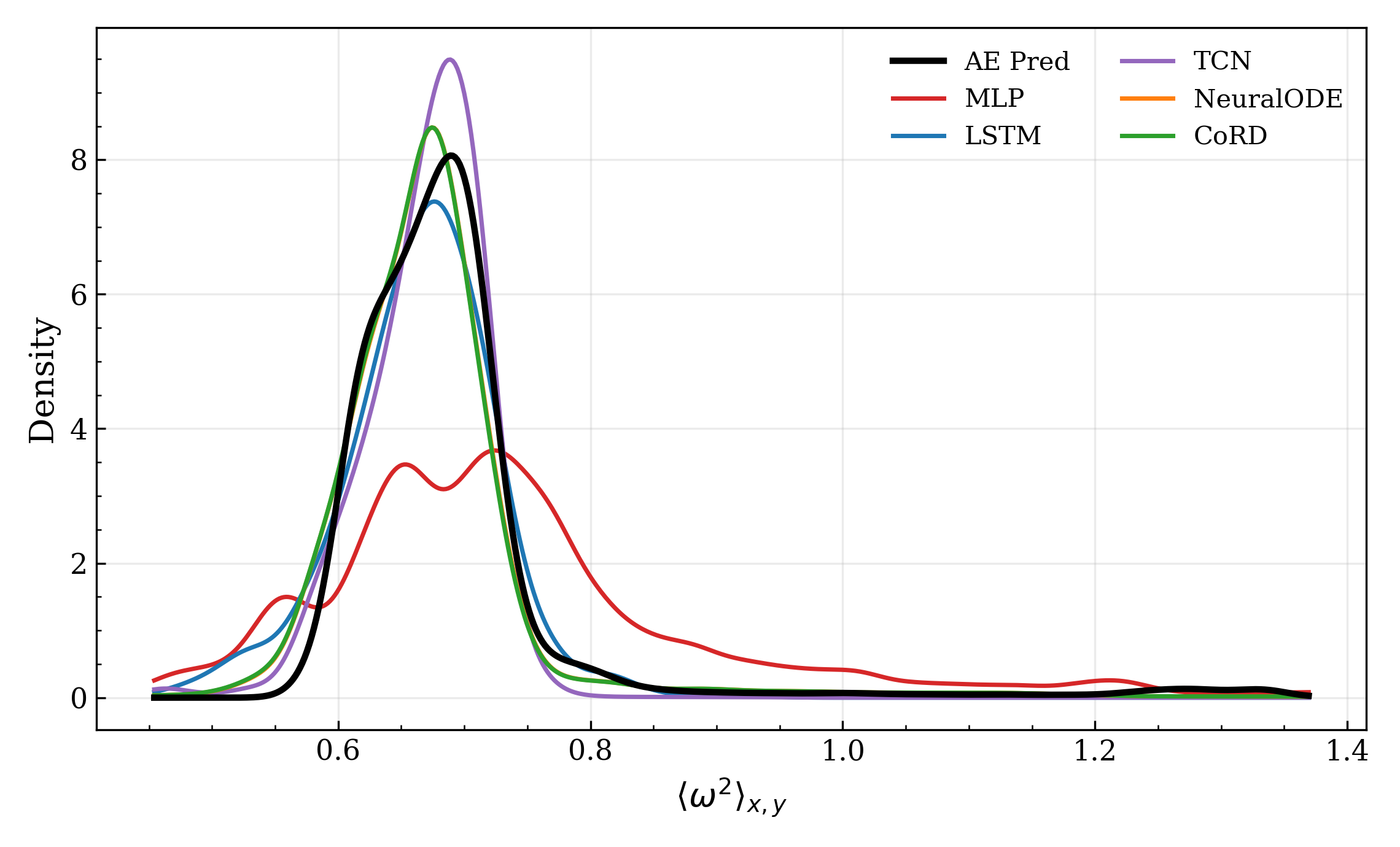}
\end{subfigure}
\vspace{0.5em}
\begin{subfigure}{0.6\linewidth}
\centering
\includegraphics[width=\linewidth]{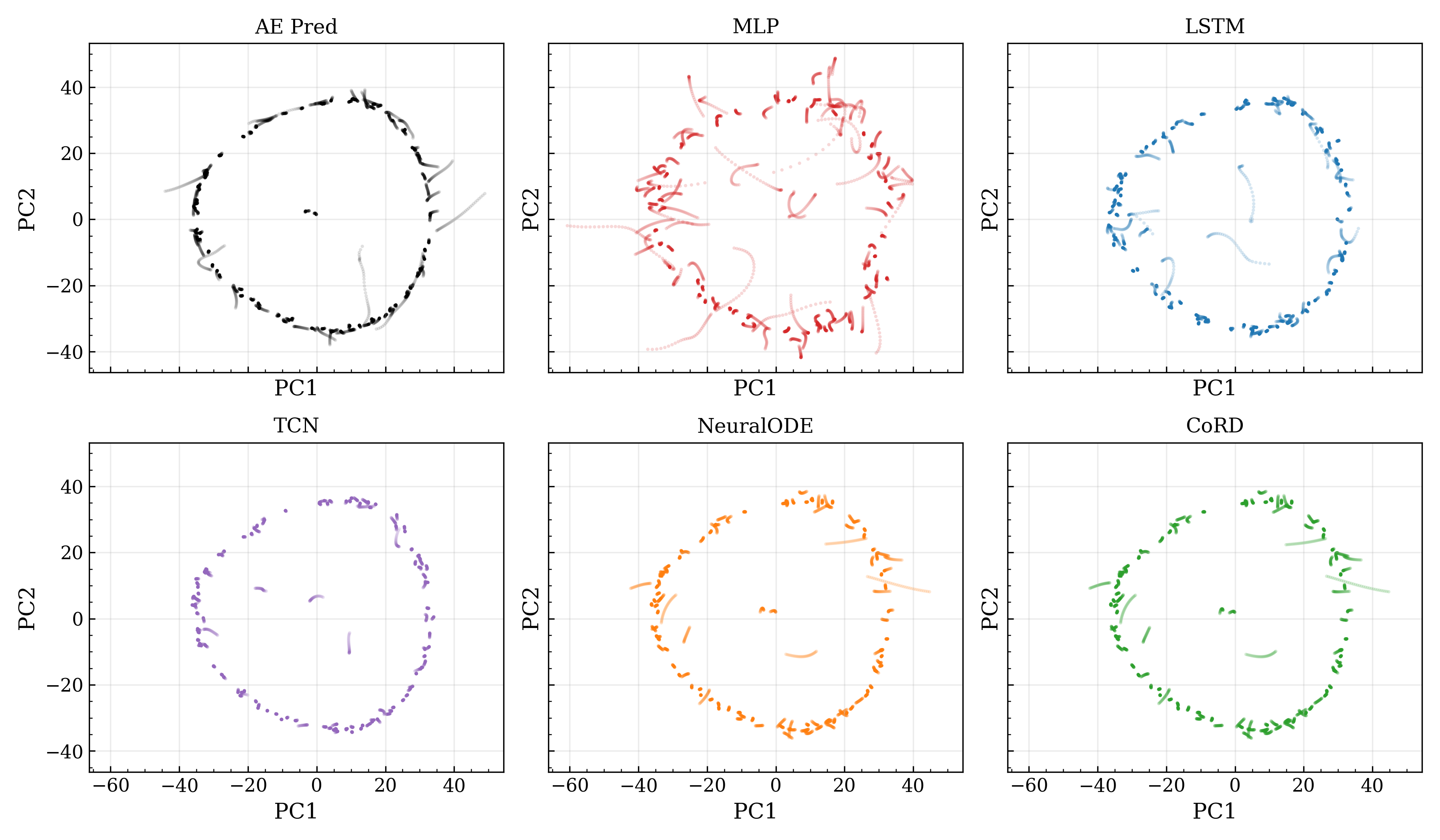}
\end{subfigure}
\caption{\textit{Kolmogorov attractor diagnostics: Top: kernel density estimates of the snapshot
enstrophy \(\langle \omega^2\rangle_{x,y}\). Bottom: PCA projections of the attractor clouds in a
basis fit on the reference AE reconstruction cloud.}}
\label{fig:kf_attractor}
\end{figure}
For Kolmogorov flow, Figure~\ref{fig:kf_attractor} shows the corresponding enstrophy distribution \(\langle \omega^2\rangle_{x,y}\) and PCA projections. The continuous-time models again remain closest to the reference density. The MLP exhibits a heavier right tail, a consequence of the expansive Jacobian as highlighted earlier. The perturbations get amplified in each step of the MLP rollout, thereby pushing the trajectories to higher-energy regions of the state space and accumulating enstrophy. The distribution of the TCN indicates that it under explores the attractor with a narrower band compared to the reference. This is consistent with the marginally contractive Jacobian behavior. 

The attractor analysis complements the local stability diagnostics. The Jacobian, bias, and finite-time amplification results explain how errors are introduced and propagated during rollout, while the attractor diagnostics reveal their long-time consequences. A model with low bias stays close to the true attractor at each prediction step, so its long-run distribution naturally approximates the true invariant measure. A model with weak perturbation amplification does not push trajectories into energetic tails that fall outside the reference attractor's distribution. The continuous-time models satisfy both conditions simultaneously, and it is precisely this combination that allows them to preserve the invariant statistics and coarse geometry of the reference attractor most faithfully across both PDE systems.
%##################################################################################
\section{Architecture Ablation}
\label{sec:arch_ablation}
\begin{figure}[]
\centering
\begin{subfigure}{0.48\linewidth}
\centering
\includegraphics[width=\linewidth]{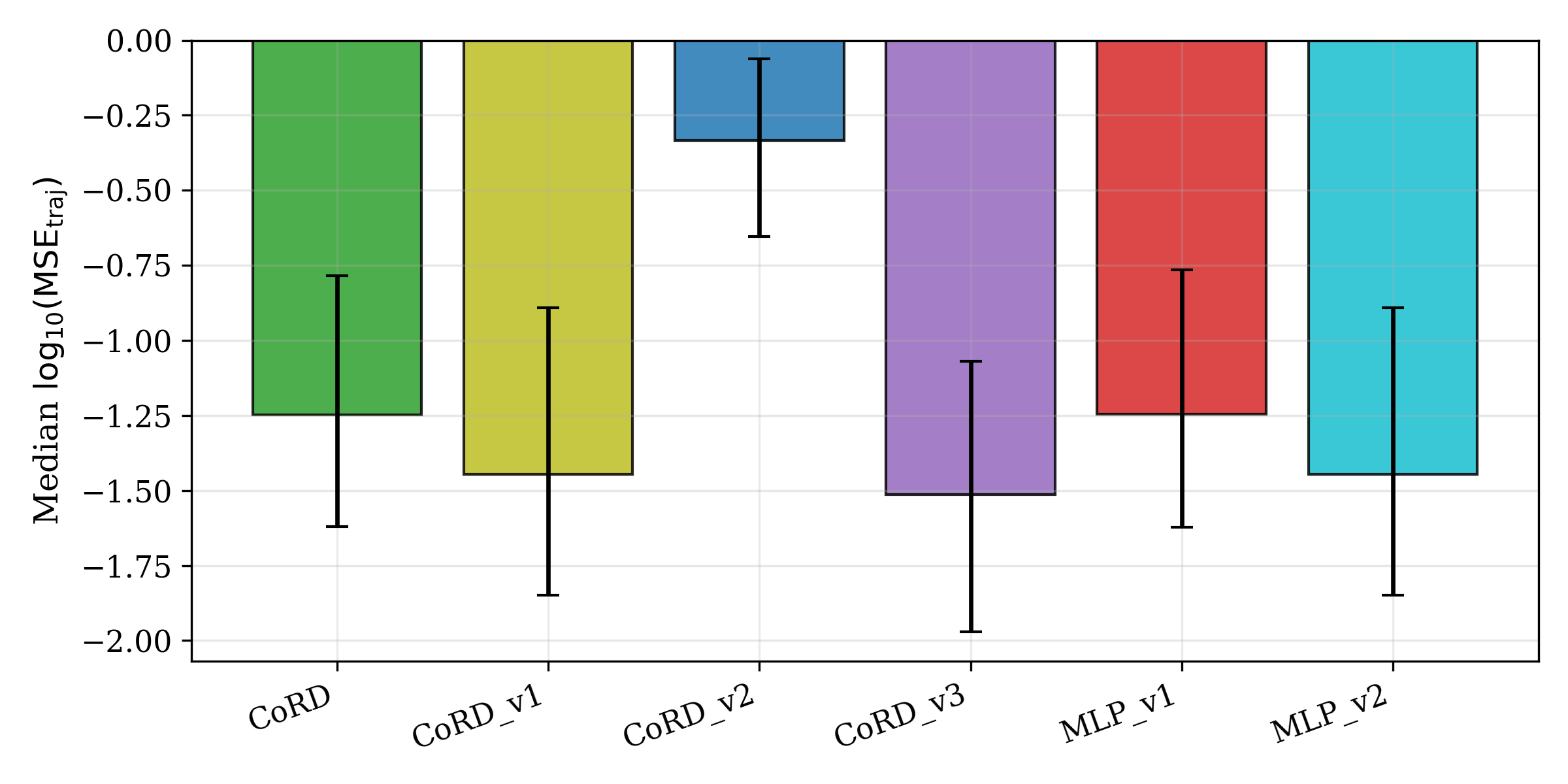}
\caption{}
\label{fig:ablation_bar}
\end{subfigure}
\hfill
\begin{subfigure}{0.48\linewidth}
\centering
\includegraphics[width=\linewidth]{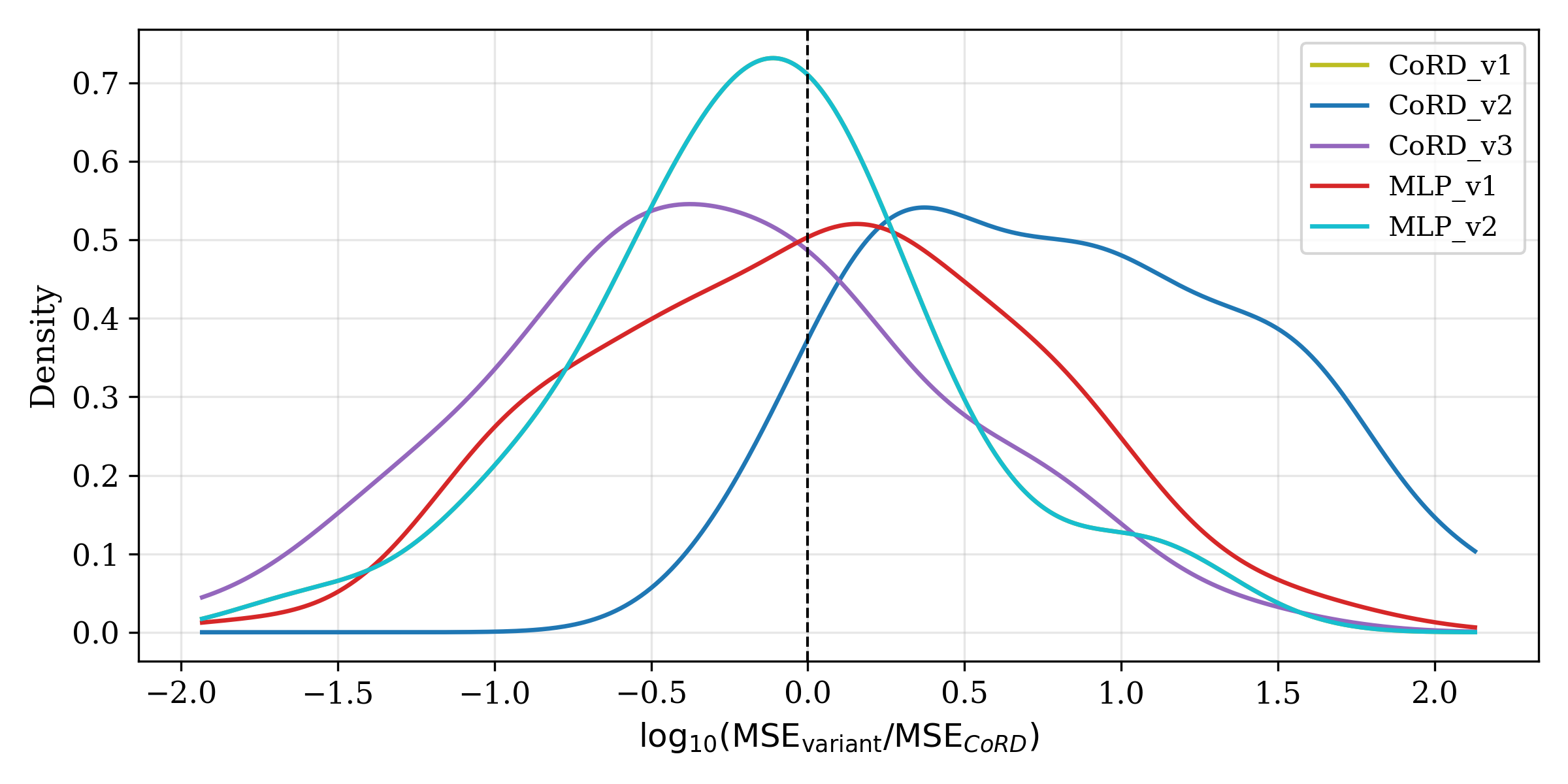}
\caption{}
\label{fig:ablation_kde}
\end{subfigure}
\caption{\textit{KS architecture ablation: Left: median trajectory-wise $\log_{10}(\mathrm{MSE}_{\mathrm{traj}})$ for each variant, with error bars denoting the interquartile range. Right: kernel density estimates of $\log_{10}(\mathrm{MSE}_{\mathrm{variant}}/\mathrm{MSE}_{\mathrm{CoRD}})$.}}
\label{fig:ablation_ks}
\end{figure}

%##################################################################################
The previous sections established the continuous-time models' capability to closely approximate complex chaotic systems. For the CoRD architecture, several design choices such as residual updates, temporal sub-stepping, and global conditioning, are bundled together. This section presents a deep dive into which design choice contributes the most for the accurate rollout performance. For this, an ablation study is carried out on the KS Equation system. Several model variants are defined as outline in Table~\ref{tab:ablation_models}.
\begin{table}[!ht]
\scriptsize
\centering
\begin{tabular}{lll}
\toprule
Model & Modification & Description \\
\midrule
CoRD & -- & Integrator model as described in Section~\ref{subsec:autoreg_CoRD}. \\
CoRD\_v1 & No sub-stepping &
Single latent update per time step ($\texttt{steps\_per\_t}=1$). \\
CoRD\_v2 & No residual update &
Directly predicts $z_{t+1}$ instead of using a residual update. \\
CoRD\_v3 & No global conditioning &
Dynamics depend only on the current latent state $z_t$. \\
MLP\_v1 & Plain autoregressive MLP &
$z_t \mapsto z_{t+1}$ without conditioning or residual structure. \\
MLP\_v2 & Residual MLP &
$z_{t+1}=z_t+\Delta t\,g(z_t,c)$ without temporal sub-stepping. \\
\bottomrule
\end{tabular}
\caption{\textit{Description of architectural variants used in the ablation study.}}
\label{tab:ablation_models}
\end{table}
Figure~\ref{fig:ablation_bar} reports the median trajectory-wise $\log_{10}(\mathrm{MSE}_{\mathrm{traj}})$ together with interquartile ranges, while Figure~\ref{fig:ablation_kde} shows the distribution of $\log_{10}(\mathrm{MSE}_{\mathrm{variant}}/\mathrm{MSE}_{\mathrm{CoRD}})$ relative to
the full model.
The clearest finding is that residual updates matter most. Removing them ($\mathrm{CoRD\_v2}$) produces the largest single degradation with a substantially higher median error and a distribution that concentrates firmly in the positive log-ratio range. This indicates that the variant is worse than the full model on the overwhelming majority of trajectories. This is not surprising as residual formulations keep each update small and close to the identity, which tends to stabilize training and prevent individual steps from over-correcting. In a chaotic system where small per-step errors compound quickly, that stabilizing effect has a significant influence on long-horizon behavior. Removing temporal sub-stepping ($\mathrm{CoRD\_v1)}$) produces a more moderate degradation, the distribution shifts slightly toward positive log-ratios but with less severity than ($\mathrm{CoRD\_v2}$). Sub-stepping allows the model to internally resolve the dynamics at a finer timescale than the output frequency, which matters for systems like KS where the relevant nonlinear interactions happen faster than the observation interval. Without it, the model has to learn a larger single-step map, which is a harder function to approximate accurately.
Eliminating global conditioning ($\mathrm{CoRD\_v3}$) also increases error and broadens the distribution considerably. The initial latent state acts as a long-horizon anchor giving the model a persistent reference to the starting conditions that doesn't decay through the recurrent state the way short-term information does. Losing that anchor appears to reduce the model's ability to maintain coherent trajectories over long rollouts, even if individual steps remain reasonable. The MLP baselines put the CoRD results in perspective. The plain autoregressive MLP is consistently worse than all CoRD variants, and the residual MLP confirms that the residual structure itself carries real value, although it remains clearly different from CoRD. The remaining gap between the residual MLP and CoRD is therefore attributable to sub-stepping and global conditioning, the two components that have no analogue in the MLP family.
%##################################################################################
\section{Discussion}
\label{sec:discussion}
%##################################################################################
In this study, a controlled comparison of five temporal architectures under a uniform autoregressive training protocol is presented. The models are categorized into two main labels, discrete-time update (MLP, LSTM, TCN) and continuous-time update (NeuralODE and CoRD). The objective is not to identify a single best architecture, rather to understand how architectural choices behave in long-horizon rollout configuration via systematic analysis. A common trend is observed across the chosen benchmarks with the continuous-time update style architectures producing more stable and accurate closed-loop rollouts than discrete-time update approaches. While this finding itself is not novel or surprising, the diagnostic analysis of the learning mechanisms certainly produces specific insights.

The error injection and propagation diagnostics quantify the differences in the models at multiple levels. The Jacobian analysis shows that the models operate in different local sensitivity regimes, namely contractive, expansive, or near-marginal. It also connects the impact of this with the long-horizon predictive accuracies. In addition, one-step bias measurement shows that continuous-time models introduce smaller errors per step, whereas FTLE captures the amplification of these errors. A key finding is that the LSTM's contractive Jacobian under teacher forcing does not preclude large perturbation growth under closed-loop rollout, because trajectory drift places the model in regions of state space with different local dynamics. This indicates that the two metrics are complementary to each other and could lead to erroneous characterization if taken in isolation. The attractor analysis complements the above by capturing the models' abilities to learn the underlying geometry of the chaotic systems. It also cements the findings from the Jacobian study and shows the states into which the trajectories evolve.

The architecture ablation study provides additional context for these observations. By systematically removing residual updates, temporal sub-stepping, and global conditioning, the ablation isolates how individual update mechanisms
affect rollout behavior. Residual updates shows the highest impact and is consistent with the findings in the dynamical analysis section. Temporal sub-stepping and global conditioning also produces consistent impact in the predictive accuracies. This section therefore presents important design guidelines for neural architectures.

Several limitations should be noted. First, the PDE experiments depend on a fixed autoencoder representation, so the conclusions concern temporal evolution in a learned latent space and as such the impact of the reconstruction inaccuracies cannot be fully rejected. Second, all models are trained under a common one-step teacher-forcing objective. This is a deliberate design choice to ensure fair comparison, but alternative training schemes could modify the balance between bias injection and long-horizon stability. Finally, none of the architectures incorporate explicit physical constraints, so the results should be interpreted as a comparison of temporal update mechanisms rather than of physics-informed surrogate models. These limitations also suggest several directions for future work. For example, combining the diagnostic framework with physics-informed or operator-learning architectures, could demonstrate how explicit inductive biases impact the error propagation mechanism outlined here.
%##################################################################################
\section{Conclusion}
\label{sec:conclusion}
%##################################################################################
This work presents a controlled comparison of five commonly used temporal architectures, namely, MLP, LSTM, TCN, NeuralODE, and CoRD for modeling chaotic dynamical systems. All models are trained and evaluated using a homogeneous protocol on benchmark problems with increasing complexity.

The central finding is that long-horizon stability is governed primarily by how architectures inject and propagate prediction error through time rather than the short-term accuracies. Local Jacobian, one-step bias, and finite-time perturbation growth provide quantification at different levels and explain the gap between the learning mechanisms of the discrete and continuous-time update models. Attractor analysis shows the final state of the learned trajectories and is a geometric representation of the surrogates' accuracies. The ablation studies further illustrate the residual-update being a core contributor to the long-horizon prediction capability.

These results suggest that assessing chaotic surrogates via trajectory error metrics only is insufficient. Dynamical diagnostics probing error propagation and invariant structure provide a more complete and mechanistically interpretable basis for comparing temporal architectures. Such should be considered alongside standard accuracy metrics for future surrogate modeling studies.
%##################################################################################
\section*{Acknowledgments}
%##################################################################################
The author received no financial support from any funding agency in the public, commercial, or not-for-profit sector for the research, authorship, and/or publication of this article.
%##################################################################################
% \section*{Declaration of Competing Interest}
% The author declares that there is no known competing interest that could have influenced the work presented in this paper.
% %##################################################################################
% %% The Appendices part is started with the command \appendix;
% %% appendix sections are then done as normal sections
% \section*{Declaration of generative AI and AI-assisted technologies in the manuscript preparation process}
% During the preparation of this work the author(s) used OpenAI's ChatGPT suite of models in order to assist in the code development phase particularly for data generation. After using this tool/service, the author(s) reviewed and edited the content as needed and take(s) full responsibility for the content of the published article.
\appendix
\renewcommand{\thesection}{Appendix \Alph{section}}
\section{MLP Architecture Details}
\label{app:mlp_arch}
For completeness, the multilayer perceptron used in the autoregressive baseline is written explicitly as a feed-forward network with $L$ layers. Given the conditioned input $\mathbf{u}_t$ defined in Section~\ref{subsec:problem_formulation},
\[
\mathbf{u}_t = [\tilde{\mathbf{x}}_t,\; \tilde{\mathbf{x}}_0],
\]
the hidden layers are computed as
\begin{align}
  \mathbf{h}^{(1)} &= \phi(\mathbf{W}^{(1)} \mathbf{u}_t + \mathbf{b}^{(1)}), \\
  \mathbf{h}^{(\ell)} &= \phi(\mathbf{W}^{(\ell)} \mathbf{h}^{(\ell-1)} + \mathbf{b}^{(\ell)}),
  \quad \ell = 2,\dots,L-1 ,
\end{align}
where $\phi(\cdot)$ denotes the GELU activation function.
The output layer produces the next-state prediction
\begin{equation}
  \tilde{\mathbf{x}}_{t+1,\text{pred}}
  = \mathbf{W}^{(L)} \mathbf{h}^{(L-1)} + \mathbf{b}^{(L)} .
\end{equation}
All layers are fully connected. Network width and depth are chosen to match the parameter budgets used in the capacity-matched comparisons across architectures.
%============================================
\section{LSTM Architecture Details}
\label{app:lstm_arch}
%============================================
For completeness, we summarize the internal gating operations of the LSTM used in the autoregressive baseline. Given input $\mathbf{u}_t$, hidden state $\mathbf{h}_t$, and cell state $\mathbf{c}_t$, the gate updates are
\begin{align}
\mathbf{i}_t &= \sigma(\mathbf{W}_i \mathbf{u}_t + \mathbf{U}_i \mathbf{h}_t + \mathbf{b}_i), \\
\mathbf{f}_t &= \sigma(\mathbf{W}_f \mathbf{u}_t + \mathbf{U}_f \mathbf{h}_t + \mathbf{b}_f), \\
\mathbf{o}_t &= \sigma(\mathbf{W}_o \mathbf{u}_t + \mathbf{U}_o \mathbf{h}_t + \mathbf{b}_o), \\
\tilde{\mathbf{c}}_t &= \tanh(\mathbf{W}_c \mathbf{u}_t + \mathbf{U}_c \mathbf{h}_t + \mathbf{b}_c).
\end{align}
The cell and hidden states are then updated according to
\begin{align}
\mathbf{c}_{t+1} &= \mathbf{f}_t \odot \mathbf{c}_t + \mathbf{i}_t \odot \tilde{\mathbf{c}}_t, \\
\mathbf{h}_{t+1} &= \mathbf{o}_t \odot \tanh(\mathbf{c}_{t+1}),
\end{align}
where $\sigma(\cdot)$ denotes the sigmoid function and $\odot$ denotes elementwise multiplication.
Multiple stacked LSTM layers are used in the experiments, with the hidden state of one layer serving as the input to the next.
%============================================
\section{TCN Architecture Details}
\label{app:tcn_arch}
%============================================
The temporal convolutional network is constructed from residual blocks composed of two dilated causal convolutions. For a hidden sequence $\mathbf{h}$, a residual block with dilation $d$ and kernel size $k$ computes
\begin{align}
\mathbf{z}^{(1)}_t &= \phi(\mathrm{Conv}_{d,k}(\mathbf{h})_t), \\
\mathbf{z}^{(2)}_t &= \phi(\mathrm{Conv}_{d,k}(\mathbf{z}^{(1)})_t), \\
\mathbf{h}'_t &= \mathbf{h}_t + \mathbf{z}^{(2)}_t .
\end{align}
Here $\phi(\cdot)$ denotes the GELU activation function. The convolutions are causal so that outputs depend only on past inputs.
Multiple residual blocks are stacked with exponentially increasing dilation factors, allowing the receptive field of the network to expand across the temporal dimension.
%============================================
\section{NeuralODE Architecture Details}
\label{app:neuralode_arch}
%============================================
The NeuralODE baseline represents temporal evolution as a continuous-time vector field
\[
\frac{d\tilde{\mathbf{x}}(s)}{ds}
= f_\theta([\tilde{\mathbf{x}}(s),\tilde{\mathbf{x}}_0]).
\]
The vector field $f_\theta$ is parameterized by a multilayer perceptron identical to the network used in the CoRD model (Section~\ref{app:cord_arch}). This ensures comparable representational capacity between the continuous-time and discrete-time formulations.
Given an initial state $\tilde{\mathbf{x}}(0)=\tilde{\mathbf{x}}_t$, the next prediction is obtained by integrating the ODE over a fixed time interval $h_{\text{ODE}}$,
\[
\tilde{\mathbf{x}}_{t+1,\text{pred}}
= \tilde{\mathbf{x}}(h_{\text{ODE}}).
\]
Numerical integration is performed using an adaptive Dormand–Prince (dopri5) solver with specified relative and absolute tolerances.
%============================================
\section{CoRD Architecture Details}
\label{app:cord_arch}
%============================================
The vector field $f_\theta$ used in the CoRD residual update is parameterized by a multilayer perceptron
\begin{align}
  \mathbf{h}^{(1)} &= \phi(\mathbf{W}^{(1)} \mathbf{u}_t + \mathbf{b}^{(1)}), \\
  \mathbf{h}^{(\ell)} &= \phi(\mathbf{W}^{(\ell)} \mathbf{h}^{(\ell-1)} + \mathbf{b}^{(\ell)}),
  \quad \ell = 2,\dots,L_{\text{dyn}}-1, \\
  f_\theta(\mathbf{u}_t)
  &= \mathbf{W}^{(L_{\text{dyn}})} \mathbf{h}^{(L_{\text{dyn}}-1)} + \mathbf{b}^{(L_{\text{dyn}})} .
\end{align}
When $K$ sub-steps are used within each observation interval, the residual update becomes
\begin{align}
  \tilde{\mathbf{x}}^{(0)} &= \tilde{\mathbf{x}}_t, \\
  \tilde{\mathbf{x}}^{(k+1)} &=
    \tilde{\mathbf{x}}^{(k)}
    + \Delta t_{\text{sub}}
      f_\theta([\tilde{\mathbf{x}}^{(k)},\tilde{\mathbf{x}}_0]),
  \quad k=0,\dots,K-1, \\
  \tilde{\mathbf{x}}_{t+1,\text{pred}} &= \tilde{\mathbf{x}}^{(K)},
\end{align}
with $\Delta t_{\text{sub}} = \Delta t/K$. This corresponds to integrating the learned vector field using multiple explicit Euler steps within each observation interval.

\bibliographystyle{plainnat}

\bibliography{cas-refs}

\end{document}